%% file: sn-article.tex
\documentclass[pdflatex,sn-nature]{sn-jnl}

\usepackage{graphicx}%
\usepackage{multirow}%
\usepackage{amsmath,amssymb,amsfonts}%
\usepackage{amsthm}%
\usepackage{mathrsfs}%
\usepackage{mathtools}%
\usepackage[title]{appendix}%
\usepackage{xcolor}%
\usepackage{textcomp}%
\usepackage{manyfoot}%
\usepackage{booktabs}%
\usepackage{algorithm}%
\usepackage{algorithmicx}%
\usepackage{algpseudocode}%
\usepackage{listings}%
\usepackage{nicefrac}%

\expandafter\def\csname [email protected]\endcsname{}
\usepackage{cleveref}%
\hypersetup{hypertexnames=false}

\crefname{figure}{Fig.}{Figs.}
\Crefname{figure}{Figure}{Figures}

\crefname{table}{Tab.}{Tabs.}
\Crefname{table}{Table}{Tables}

\crefname{extfigure}{Extended Data Fig.}{Extended Data Figs.}
\Crefname{extfigure}{Extended Data Figure}{Extended Data Figures}

\crefname{exttable}{Extended Data Tab.}{Extended Data Tabs.}
\Crefname{exttable}{Extended Data Table}{Extended Data Tables}

\usepackage{multirow}
\usepackage{makecell}
\usepackage{minitoc}

\usepackage{multibib}
\newcites{methods}{Methods References}

\theoremstyle{thmstyleone}%
\theoremstyle{thmstyletwo}%
\theoremstyle{thmstylethree}%

\input{macros}

\makeatletter
\renewcommand{\email}[1]{%
  \global\advance\emailcnt by 1\relax%
  \if@corauemail%
    \g@addto@macro\corrauthemail{%
      \setcounter{footnote}{0}%
      \textcolor{black}{#1};\ %
    }%
  \else%
    \g@addto@macro\authemail{%
      \setcounter{footnote}{0}%
      \textcolor{black}{#1};\ %
    }%
  \fi
}
\makeatother

\begin{document}

\title[Quanta Perception]{\vspace{-0.5in}Quanta Perception as Probabilistic Events}

\author*[1]{\fnm{Varun} \sur{Sundar}}\email{vsundar4@wisc.edu}

\author[1]{\fnm{Pavan} \sur{Thodima}}\email{thodima@wisc.edu}

\author[1]{\fnm{Sacha} \sur{Jungerman}}\email{sjungerman@wisc.edu}

\author[1]{\fnm{Mohit} \sur{Gupta}}\email{mohitg@cs.wisc.edu}

\makeatletter
\g@addto@macro\authemail{\par\href{https://wisionlab.com/project/probabilistic-events/}{\texttt{wisionlab.com/project/probabilistic-events/}}.}
\makeatother

\affil[1]{\orgdiv{Department of Computer Sciences}, \orgname{University of Wisconsin--Madison}, \orgaddress{\state{Wisconsin}, \country{United States}}}

\input{sections/abstract}
\maketitle

\input{sections/introduction}
\clearpage
\input{sections/results}
\input{sections/discussion}

\clearpage
\bibliography{references}
\clearpage

\input{sections/methods}

\clearpage
\bibliographystylemethods{sn-nature}
\bibliographymethods{references}
\clearpage

\section*{Declarations}

\subsection*{Funding}

This research was supported in parts by an NSF CAREER award, a SONY Faculty Innovation Award, and a WARF Research Forward Initiative Award.

\subsection*{Competing Interests}

The authors declare no competing interests.

\subsection*{Ethics Approval and Consent to Participate}

Not applicable. This study does not involve experiments on human or animal subjects.

\subsection*{Consent for Publication}
Not applicable. The images in this manuscript do not contain identifiable human data.

\subsection*{Data Availability}

The quanta sensor data supporting \cref{fig:method_walkthrough,fig:plug_and_play,fig:qbp_comparison,fig:comparison_to_conventional_cameras,fig:noise_blur,fig:reconstruction_results,fig:spatially_varying_denoising,fig:teaser,fig:color_metering} are openly available at \url{https://github.com/wision-lab/datasets}. Data for \cref{fig:limitations} were sourced from Liu et al. \citemethods{liu2024bit2bit}, and the VisionSim dataset used for the Methods analysis is available at \url{https://visionsim.readthedocs.io/en/latest/}.

\subsection*{Materials Availability}
Not applicable.

\subsection*{Code Availability}
The code supporting this study is provided as a compressed supplementary file. The archive includes scripts for generating the quantitative tables and plots reported in the manuscript from the cited datasets.

\subsection*{Author Contribution}

\textbf{V.S.} conceived the core concept, developed the theory and software, designed experiments, and wrote the manuscript. 
\textbf{P.T.} implemented baselines and visualization tools, and acquired data. 
\textbf{S.J.} acquired data, analyzed results, and contributed to writing. 
\textbf{M.G.} conceptualized the study and experimental design, secured funding, and contributed to writing. 
All authors approved the final manuscript.

\clearpage
\section*{Extended Data}

\renewcommand{\theHtable}{ED.\arabic{table}}
\renewcommand{\theHfigure}{ED.\arabic{figure}}

\setcounter{table}{0}
\setcounter{figure}{0}

\renewcommand{\tablename}{Extended Data Table}
\renewcommand{\figurename}{Extended Data Figure}

\renewcommand{\thetable}{\arabic{table}}
\renewcommand{\thefigure}{\arabic{figure}}

\crefname{figure}{Extended Data Fig.}{Extended Data Figs.}
\Crefname{figure}{Extended Data Figure}{Extended Data Figures}

\crefname{table}{Extended Data Tab.}{Extended Data Tabs.}
\Crefname{table}{Extended Data Table}{Extended Data Tables}

\crefalias{figure}{extfigure}
\crefalias{table}{exttable}

\input{figures/reconstruction_results}

\input{tables/i2k_metrics}
\clearpage

\input{figures/reconstruction_results_qualitative}
\clearpage

\input{tables/visionsim_depth_flow_metrics}
\input{figures/visionsim_depth_flow}
\clearpage

\input{figures/noise_blur}
\input{figures/using_spatial_info}
\input{figures/metering_runtime}
\input{figures/spatially_varying_denoising}
\input{figures/rate_distortion_events_vs_video_codec}
\input{figures/color_metering}
\input{figures/analysis_exposure_time}

\input{figures/analysis_summed_frames_vs_light_level}
\input{figures/analysis_summed_frames_vs_light_level_aggregate}

\input{figures/analysis}

\end{document}

%% file: macros.tex
\newcommand{\vp}{\mathbf{p}}

\newcommand{\rr}{\mathbb{R}}

\usepackage{bm}

\DeclarePairedDelimiter{\paren}{(}{)}
\DeclarePairedDelimiter{\curly}{\{}{\}}

\newcommand{\Prob}[1]{P\paren*{#1}}

\newcommand{\tildeNice}{{\raise.17ex\hbox{$\scriptstyle\sim$}}}
\newcommand{\greaterNice}{{\raise.17ex\hbox{$\scriptstyle >$}}}
\newcommand{\approxNice}{{\raise.15ex\hbox{$\scriptstyle\approx$}}}
\newcommand{\ggNice}{{\raise.15ex\hbox{$\scriptstyle >> $}}}

\usepackage{newpxtext}

\ExplSyntaxOn

\NewDocumentCommand { \ie } { }
  {
    \textit{i.e.}
    \peek_meaning_ignore_spaces:NTF .
      { \skip_horizontal:n { -.3ex } \use_none:n }
      {
        \peek_meaning_ignore_spaces:NF ,
          { \skip_horizontal:n { -.3ex } }
      }
  }
\NewDocumentCommand { \eg } { }
  {
    \textit{e.g.}
    \peek_meaning_ignore_spaces:NTF .
      { \skip_horizontal:n { -.3ex } \use_none:n }
      {
        \peek_meaning_ignore_spaces:NF ,
          { \skip_horizontal:n { -.3ex } }
      }
  }

\NewDocumentCommand { \etc } { }
  {
    etc.
    \peek_meaning_ignore_spaces:NT .
      { \use_none:n }
  }

\NewDocumentCommand { \cad } { }
  {
    \textit{c-à-d.}
    \peek_meaning_ignore_spaces:NTF .
      { \skip_horizontal:n { -.3ex } \use_none:n }
      {
        \peek_meaning_ignore_spaces:NF ,
          { \skip_horizontal:n { -.3ex } }
      }
  }
\ExplSyntaxOff

%% file: sections/abstract.tex
\abstract{Autonomous systems rely on extracting information from light, yet remain brittle in extreme environments, from nighttime navigation to high-speed robotics. Conventional sensors aggregate photons over fixed exposures, imposing trade-offs between sensitivity, dynamic range, and temporal resolution that degrade perception when photons are scarce or dynamics are rapid. Quanta sensors detect individual photons, but their streams exceed real-time compute and latency budgets by orders of magnitude.

Here we introduce \textit{probabilistic events}, a computational primitive for real-time quanta perception from individual photon detections. By computing the posterior over the time since the last intensity change, we represent photon streams as recursive belief states. Rather than fixed-threshold event-camera triggers, this recursive Bayesian formulation yields three low-latency signals: motion-adaptive scene flux, high-fidelity activity maps, and entropy-based perceptual uncertainty. This representation enables perception in extreme conditions, including pose estimation of a running person at $\sim$0.05 lux---without retraining vision models. 
Our approach processes input streams exceeding 50{,}000 quanta frames per second on commodity GPU hardware---yielding kilohertz-scale outputs up to four orders of magnitude faster than state-of-the-art quanta reconstruction baselines, even for megapixel arrays.
By replacing frame reconstruction with direct probabilistic inference over photon streams, this work bridges photon-counting quanta sensing with robotic vision.}

\keywords{Computational imaging, quanta sensing, event cameras, robot perception, low-latency vision, Bayesian inference}

%% file: sections/introduction.tex
\section*{Introduction}

Vision is a central interface between the physical world and modern autonomous and robotic systems, yet it is often the weakest link when these systems leave controlled environments. In low light, fast motion, or scenes with extreme dynamic range, perception failures arise not just from insufficient algorithms, but from information lost at the moment of sensing. Conventional image sensors measure light by integrating photons over fixed exposure windows (\cref{fig:teaser}A), a strategy that entangles sensitivity, noise, motion blur, and dynamic range into a single measurement, discarding the continuous-time dynamics of photon arrival. As a result, autonomous systems fail most often in exactly these regimes---where robustness matters most. This sensory bottleneck poses a significant barrier to the development of ``all-weather'' physical artificial intelligence.

Quanta (photon-counting) image sensors~\citep{fossum2005sub}\citep{fossum:11} offer a fundamentally different sensing modality. They capture light at its finest granularity---the individual photon---thereby preserving the discrete arrival statistics of light (\cref{fig:teaser}B). This enables sensing across illumination levels~\citep{ma2022review,ma_quanta_2020,inglehighfluxpassive2019,inglepassiveinterphotonimaging2021} and motion dynamics~\citep{Wei_2023_ICCV,Sundar_2024_CVPR} in regimes that are beyond the capabilities of conventional imagers. After decades of development, recent advancements---particularly those based on single-photon avalanche diodes (SPADs)---have transformed the quanta sensing landscape. The emergence of megapixel-scale arrays~\citep{ma:17,ma2021photon,ma20210,morimoto_cannon,morimoto_megapixel_2020,ulku512512spad2019,jaijuiMa2022} now, for the first time, makes it feasible to perform computer vision in the wild using photon-level measurements.

\input{figures/teaser}

Paradoxically, the abundance of photon-level information captured by quanta sensors has so far prevented them from being used for real-time autonomy. The sensor emits photon detections as high-speed binary streams, typically arranged as quanta frames sampled at tens to hundreds of kilohertz. For a 1-megapixel sensor, this exceeds 100 Gbits/s of throughput, far beyond the bandwidth, compute, and latency envelopes of practical robotic systems. Yet the challenge extends beyond transmission: extracting representations compatible with downstream vision currently requires computationally intensive reconstruction algorithms~\citep{ma_quanta_2020,chennuri2025quanta,Sundar_2024_CVPR}. Such methods often require minutes to process mere seconds of acquisition---effective for offline analysis, but fundamentally misaligned with low-latency operation in dynamic scenes where photon scarcity and motion interact. The dilemma is clear: we now have sensors that capture the world at the granularity of individual photons, but we lack representations that can turn photon streams into \emph{actionable perception}---representations that summarize what has changed, over what temporal scale, and with what degree of confidence, while interfacing directly with off-the-shelf vision models under the latency and compute constraints of real robotic systems.

In this work, we address this tension by introducing \textit{probabilistic events} as a new computational primitive for quanta perception (\cref{fig:teaser}C). The key insight is to formulate visual sensing as a recursive Bayesian inference problem, tracking the probability distribution of the intensity run-lengths, defined as the time elapsed since the last intensity transition~\citep{adams2007bayesian}. (Here, \textit{event} denotes an inferred intensity change, not the fixed-threshold trigger of a dynamic vision sensor.) Each pixel recursively maintains competing hypotheses about whether recent photon arrivals are consistent with the established intensity regime or indicate a change. This generalizes standard event cameras by replacing deterministic thresholded triggers with a probabilistic belief state over change and uncertainty. From the inferred distribution, we synthesize three complementary information streams: a motion-aware photon aggregate, a temporal stability map, and significant entropy shifts reflecting changes in belief uncertainty---introducing a form of sensor-level metacognition absent in conventional imaging.

Formulating quanta perception as probabilistic events reconciles the fundamental limits of photon counting with the latency constraints of real-time autonomy. We demonstrate input throughputs exceeding 50,000 quanta frames per second (qFPS) on desktop-grade GPU hardware, with output throughputs up to four orders of magnitude faster than state-of-the-art reconstruction baselines. This translates to an output throughput in the kilohertz range, ensuring that for a wide gamut of downstream tasks, the extraction of high-fidelity representations from photon streams is no longer the bottleneck of the quanta perception pipeline. As shown in \cref{fig:teaser}D, this representation serves as a direct interface for off-the-shelf vision models without retraining, supporting low-level features, mid-level geometry, and high-level semantics. We demonstrate robust inference in conditions previously inaccessible to real-time systems, including detecting and estimating the pose of a running person at less than 0.05 lux illumination and high-fidelity geometric-semantic understanding in nighttime driving. By redefining the elementary units of vision at the photon level, probabilistic events transform quanta sensors from offline scientific imaging devices into practical, all-weather perception sensors for autonomous systems.

%% file: figures/teaser.tex
\begin{figure*}[t]
    \centering
    \includegraphics[width=\textwidth]{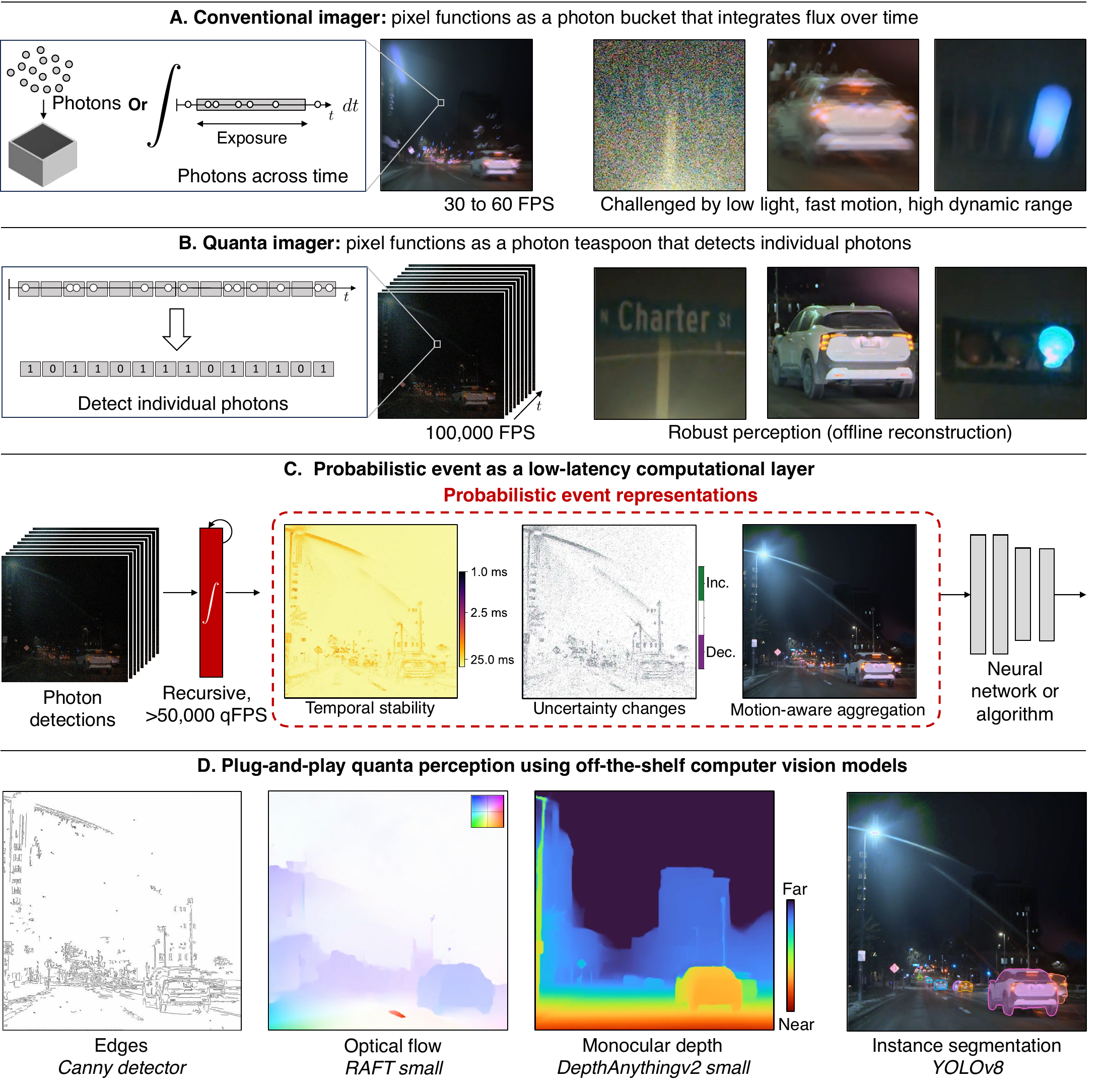}
    \vspace{-0.2in}
    \caption{\textbf{From photons to perception using probabilistic events.} 
    \textbf{(A)} Conventional image sensors function as \textit{photon buckets}, integrating flux over coarse exposure windows (tens of milliseconds): a mechanism that fundamentally limits performance in low-light and high-dynamic-range scenarios. 
    \textbf{(B)} Conversely, quanta image sensor pixels act as \textit{photon teaspoons}, registering individual photon arrivals. While this detection capability allows quanta cameras to operate across extreme illumination and motion regimes, it comes at the cost of massive data volumes. Consequently, existing reconstruction techniques remain offline methods that impose prohibitive compute, readout, and latency demands, precluding real-time perception. 
    \textbf{(C)} To bridge this gap, we introduce probabilistic event cameras, a computational sensing layer that operates directly on photon streams. Unlike standard event cameras that emit deterministic thresholded events, probabilistic events maintain a posterior distribution over change intervals, yielding representations that encode both signal and uncertainty---temporal stability, entropy shifts, and motion-aware flux aggregates. 
    \textbf{(D)} This representation enables motion-aware photon aggregation and uncertainty estimation at low latency and constant memory, transforming quanta cameras into plug-and-play sensors for \textit{extreme} computer vision. \textbf{Using real acquisition from a 1-megapixel color quanta sensor}, we demonstrate perception results for edge detection (Canny~\citep{CannyEdge1986}), optical flow estimation (RAFT small~\citep{teed2020raft}), monocular depth estimation (DepthAnything-v2-s~\citep{yang2024depth}), and instance segmentation (YOLO-v8 nano~\citep{Jocher_Ultralytics_YOLO_2023}).}
    \label{fig:teaser}
\end{figure*}

%% file: sections/results.tex
\section*{Results}\label{sec:results}

\subsection*{Probabilistic Event Representations}

Probabilistic events are a task-ready representation computed by Bayesian inference on photon streams. Rather than accumulating photons over fixed windows or reconstructing intensity frames explicitly, each pixel maintains a recursive posterior over the run length, which is the time elapsed since the last intensity transition event. As illustrated in \cref{fig:method_walkthrough}A, each probability mass of the posterior evaluates a specific hypothesis regarding the run length---effectively modeling the arrival of photons as a multiple hypothesis inference problem. This run-length distribution serves as a per-pixel belief state over temporal stability, updated in constant time as photon detections arrive. 

\input{figures/method_walkthrough}

From the run-length posterior, we derive three complementary, low-latency perceptual signals (\cref{fig:method_walkthrough}B). First, the expected time since last change forms a continuous temporal stability map: small stability values indicate recent or ongoing change, whereas large values correspond to temporally stable structure. Temporal stability maps can be interpreted as a multi-scale probabilistic activity field that encodes the temporal persistence of structure, spanning time scales from a few hundred microseconds to several seconds. Unlike binary event maps, these maps encode both the degree and timescale of stability, providing a fine-grained measure of how long structure has persisted. Crucially, this formulation allows the representation to capture smooth spatiotemporal transitions (such as gradual illumination changes or soft motion gradients), rather than being constrained to discrete, step-like intensity changes. We illustrate this dynamic behavior in \cref{fig:method_walkthrough}B by tracking the stability time-series and log-probability distribution at a single pixel, marked by a cyan cross. Before the ping pong ball crosses this location ($t=17$~ms), the posterior distribution is highly peaked at a large stability value, reflecting a static background. As the object occludes the pixel ($t=27$~ms), the distribution shifts its probability mass to hypotheses representing shorter run lengths, causing a sharp drop in expected stability. Once the object passes ($t=37$~ms), the distribution recovers, forming a new peak as stability begins to accumulate again.

Second, the change in entropy of the run-length posterior provides a per-pixel trace of uncertainty evolution. Rising entropy reflects increasing ambiguity in the inferred temporal stability, whereas falling entropy indicates a convergence toward confident estimation. This differential signal yields a measure of information dynamics that can detect failure modes or modulate attention in downstream perception systems. 
Finally, an adaptive aggregate, modulated by the inferred temporal stability, produces a motion-aware flux estimate.
By integrating photons according to inferred temporal stability, this signal suppresses shot noise in static regions while limiting motion blur in dynamic areas. Together, these three streams---temporal stability, change in entropy, and motion-aware flux---constitute the probabilistic event representation.

\cref{fig:method_walkthrough}C illustrates probabilistic event representations across diverse light and motion regimes. During rapid ego-motion, the temporal stability map contracts around edges, preserving fine spatial structure. 
In photon-starved environments, like juggling in low light, stable regions accumulate evidence over extended durations, maintaining signal fidelity despite sparse arrivals.

The formulation is statistically general. It applies to 1-bit photon measurements via Bernoulli likelihoods and extends naturally to multi-bit measurements through Binomial models. 
This multi-bit support enables robust spatial feature computation---such as spatial gradients and Log-Gabor filters---which is particularly valuable in low-photon conditions and improves pixel responsivity to abrupt changes (\cref{fig:using_spatial_info}). Crucially, while stability estimation leverages these spatial contexts, the flux aggregation it drives is performed independently per pixel. This separation provides natural pathways for integrating with mosaicked photon detections for color quanta sensing. \cref{fig:teaser,fig:color_metering} show examples of these results captured with a megapixel color quanta camera.

By performing inference directly on photon streams, probabilistic events navigate the noise--blur trade-off adaptively, without incurring the computational burden of explicit motion-compensated reconstruction. This approach thus enables bounded-state Bayesian inference directly on photon streams at quanta data rates, while producing task-ready perceptual fields. Throughout this work, we report both the maximum sustainable input throughput of the probabilistic-event pipeline (qFPS) and the resulting output throughput (FPS) of the generated representations. Operating orders of magnitude faster than existing reconstruction-based baselines (\cref{fig:method_walkthrough}D), this efficiency shifts the dominant latency from interpreting the photon stream to the downstream perception task. Because these streams are image-like and directly consumable by standard vision models, probabilistic events transform photon-counting quanta cameras into real-time perceptual front ends for vision and robotics operating in extreme regimes.

\subsection*{Low-Latency Quanta Perception in the Wild}

We next evaluate probabilistic events as a \emph{drop-in representation} for real-world robotic perception. \Cref{fig:plug_and_play} demonstrates this versatility across a hierarchy of tasks, ranging from low-level feature extraction to high-level semantic understanding, spanning classical algorithms and modern deep neural networks---without retraining, architectural modification, or task-specific tuning.

In high-speed regimes, motion-aware flux aggregates effectively freeze motion, enabling edge detection at 4{,}000~FPS, high-speed object tracking at over 550~FPS, optical flow estimation at around 150~FPS, and QR code decoding at over 150~FPS. In low-light environments, adaptive integration accumulates sparse signal without ghosting, supporting nighttime vehicle detection and human pose estimation in a dimly lit hall at hundreds to thousands of frames per second. As shown in \cref{fig:plug_and_play}, this representation supports accurate inference in dynamic environments that typically confound conventional cameras---resolving the swinging spheres of a Newton's cradle, decoding a QR code mid-flight, and tracking a runner without succumbing to motion blur.
The end-to-end throughputs reported in \cref{fig:plug_and_play}, corresponding to end-to-end perception throughput that includes both probabilistic-event generation and downstream task execution, demonstrate that extracting perceptual representations from photon streams is no longer the dominant bottleneck in the perception pipeline.
Indeed, for extremely lightweight downstream tasks (\eg, edge detection), we use a faster variant of probabilistic events capable of up to 9,000~FPS at this resolution, fully exploiting the underlying speed of the algorithms.

A distinctive property of motion-aware aggregation is heteroscedastic noise: dynamic regions integrate fewer photons and are therefore noisier than static backgrounds. In practice, both classical operators and modern neural networks remain robust to this spatially varying uncertainty, enabling accurate inference directly from the probabilistic aggregates. For visualization only, a lightweight spatially varying denoiser (\eg, Wiener filtering) can be applied (\cref{fig:spatially_varying_denoising}).

To quantify task performance under controlled ground truth, we further evaluate downstream perception on the VisionSim dataset~\citep{visionsim}, where precise depth and flow references are available (tabulated in \cref{tab:visionsim_depth_flow_metrics}, qualitative examples shown in \cref{fig:visionsim_depth_flow}). We compare three ways of presenting quanta measurements to standard vision models: fixed-window virtual exposures, reconstruction-first perception using quanta image restoration methods (QBP~\citep{ma_quanta_2020} and QUIVER~\citep{chennuri2025quanta}), and probabilistic events using the Log-Gabor variant. The same pretrained networks are then applied to each representation, using DepthAnything-v2~\citep{yang2024depth} for monocular depth and RAFT~\citep{teed2020raft} for optical flow. Across photon budgets and motion extents, probabilistic events improve substantially over fixed-window exposures and remain competitive with reconstruction-first perception, while avoiding the latency of full image restoration. This supports the use of probabilistic events as a task-ready representation for perception, rather than only as a fast approximation to image reconstruction.
\input{figures/plug_and_play}

\clearpage

\subsection*{Comparison with Offline Quanta Image Reconstruction}

The preceding results show that probabilistic events can serve as a plug-and-play input to standard vision models at low latency. A natural next question is how this representation compares to the dominant reconstruction-first paradigm in quanta imaging, where photon streams are converted into high-fidelity intensity images before perception~\citep{ma_quanta_2020,chennuri2025quanta,Tassano_2020_CVPR,zhang2024streaming}. These methods can yield visually compelling imagery in photon-starved regimes, but typically rely on dense alignment or deep restoration pipelines, incurring computational costs that are incompatible with tight latency budgets in robotics.

We compare probabilistic events with Quanta Burst Photography (QBP)~\citep{ma_quanta_2020}, a representative offline method that performs patch-level alignment of binary frames to recover high-quality imagery under extreme shot noise. \Cref{fig:qbp_comparison} shows challenging low-light dynamic scenes with fewer than 0.05 photons per pixel per frame, where measurements are dominated by shot noise and motion. In this regime, naïve photon summation fails due to the noise--blur trade-off: increasing integration suppresses noise but introduces blur, while shortening exposure preserves motion but yields insufficient signal. QBP mitigates this trade-off through explicit motion compensation and patch-level alignment, producing visually cleaner reconstructions. However, this dense spatio-temporal alignment incurs substantial computational cost and latency.

\input{figures/qbp_comparison}

Probabilistic events take a fundamentally different approach where instead of aligning frames, each pixel maintains a posterior over temporal stability and adaptively integrates photons based on the inferred run-length distribution. The resulting motion-aware aggregation can be interpreted as a reconstruction, but one computed with bounded memory and constant-time updates per pixel. In practice, this enables high-throughput, causal processing while allowing temporally stable regions to integrate evidence over very long horizons ($10^4$--$10^5$ binary frames) without blurring dynamic structure. 
As shown in \cref{fig:qbp_comparison}, this lightweight representation sustains robust perception under the same extreme conditions while operating at orders-of-magnitude lower latency---up to four orders higher throughput than reconstruction-first quanta restoration methods (see \cref{fig:method_walkthrough}D and \cref{tab:i2k_metrics,fig:metering_runtime}).

We further compare with learned quanta restoration baselines (\eg, QUIVER~\citep{chennuri2025quanta}, gQIR~\citep{garg2026gqir}, NAFNet~\citep{chen2022simple}, FastDVDNet~\citep{Tassano_2020_CVPR}, and streaming EMA~\citep{zhang2024streaming}), report quantitative image-quality metrics on the i2k high-speed dataset~\citep{chennuri2025quanta}, and summarize throughput across implementations (see \cref{fig:reconstruction_results,tab:i2k_metrics,fig:reconstruction_results_qualitative}). These results highlight a key distinction: methods optimized for photorealistic restoration prioritize visual fidelity at high computational cost, whereas the proposed task-ready perceptual representations can be produced causally from photon streams under extremely tight latency budgets.

\subsection*{Comparison with Conventional Low-Light and High-Speed Imaging}

\input{figures/comparison_to_conventional_cameras}

We next position probabilistic events with respect to commercial cameras optimized for specific challenging operating regimes, including a standard DSLR (Canon EOS 77D), a high-speed camera (Photron Infinicam), a dedicated low-light security camera (Bosch DINION IP starlight 8000 MP), and a commercial event camera (Prophesee EVK4). These devices differ substantially in optics, pixel geometry, and on-board processing, and are each engineered to address a particular sensing constraint. Our goal is therefore not a hardware benchmark, but to assess whether a single photon-counting sensor, coupled with the proposed computational layer, can produce image-like information streams that remain usable across regimes typically covered by multiple specialized devices.

As shown in \cref{fig:comparison_to_conventional_cameras}, traditional sensing modalities suffer from various failure modes. We first consider photon-limited regimes where conventional intensity cameras face a fundamental noise--blur trade-off. In a nighttime city scene, the DSLR operating at 30~FPS captures insufficient signal, producing severely underexposed images that preclude downstream object detection. Conversely, in a dim scene with fast motion (a rapidly moving QR code), the high-speed camera must operate at high frame rates (\eg, 200~FPS) to freeze motion, but the resulting short exposures become photon-starved and yield a dark, low-contrast image. In a low-light scene with substantial ego-motion, a dedicated low-light Bosch DINION security camera achieves higher brightness by relying on long exposures. While it avoids catastrophic signal loss, it still incurs noticeable motion blur that degrades fine structure. In all cases, probabilistic events adapt integration to local temporal stability, yielding an adaptive flux field that supports successful vehicle detection in the nighttime city scene, QR decoding under high-speed motion, and robust depth recovery using an off-the-shelf modern monocular depth estimation architecture~\citep{yang2024depth} even in low-light and strong motion. 

Finally, we compare with a commercial event camera. Polarity events are highly efficient for capturing fast changes, but they discard absolute intensity information needed for recovering scene semantics. As a result, image reconstructions produced from the Prophesee EVK4 using E2VID~\citep{rebecq19pami} exhibit missing texture and artifacts. In contrast, probabilistic events retain richer temporal change information and an image-like flux field, supporting semantic models such as Segment Anything~\citep{Kirillov_2023_ICCV} that rely on texture and intensity cues.

%% file: figures/method_walkthrough.tex
\begin{figure*}[t]
    \centering
    \includegraphics[width=0.95\textwidth]{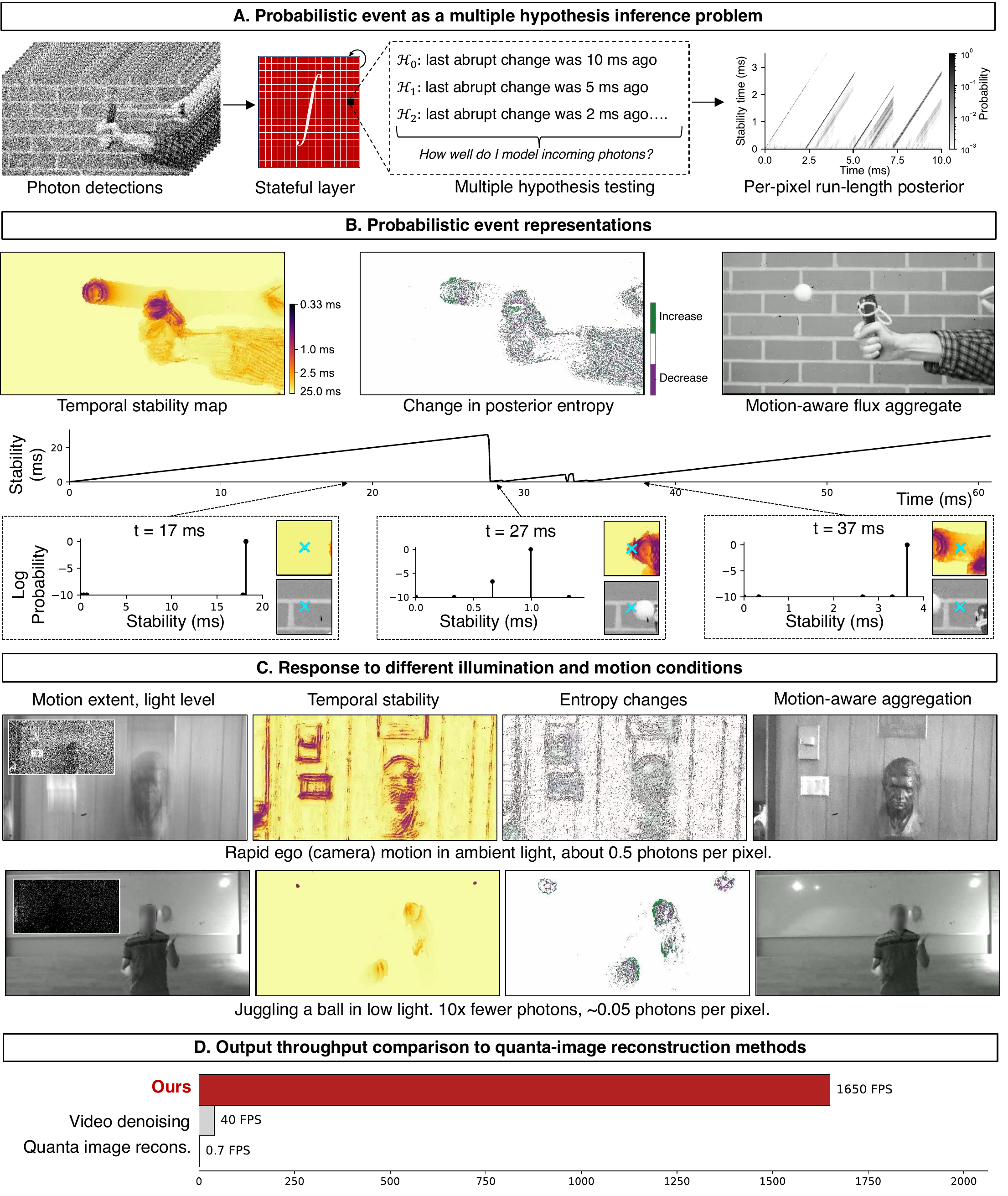}
    \vspace{-0.1in}
\caption{\textbf{Probabilistic event cameras produce low-latency, uncertainty-aware representations.}
\textbf{(A)} We treat photon arrival as multiple hypothesis inference, evaluating run length: the time since the last abrupt change.
\textbf{(B)} From the run-length posterior, we derive entropy shifts and per-pixel temporal stability maps. The latter drives a motion-aware flux aggregate that balances the noise--blur trade-off. Tracking a specific pixel (cyan cross) shows the posterior peaking at high stability before a ping pong ball crosses (t=17 ms), spreading across recent changes during occlusion (t=27 ms), and regaining a sharp peak afterwards (t=37 ms).
\textbf{(C)} Entropy fluctuations (green: increase, purple: decrease) reveal subtle ego-motion; short integration preserves high-frequency details (snapping tape); and our model robustly tracks objects despite extreme photon sparsity (low-light juggling). \textbf{We blur faces for privacy.}
\textbf{(D)} Sustaining an input throughput of 51~kqFPS and an output throughput of 1650~FPS at 256×512, our approach operates orders of magnitude faster than reconstruction and denoising baselines, enabling real-time downstream perception (\cref{fig:plug_and_play}).
}
    \label{fig:method_walkthrough}
\end{figure*}

%% file: figures/plug_and_play.tex
\begin{figure*}[t]
    \centering
    \includegraphics[width=\textwidth]{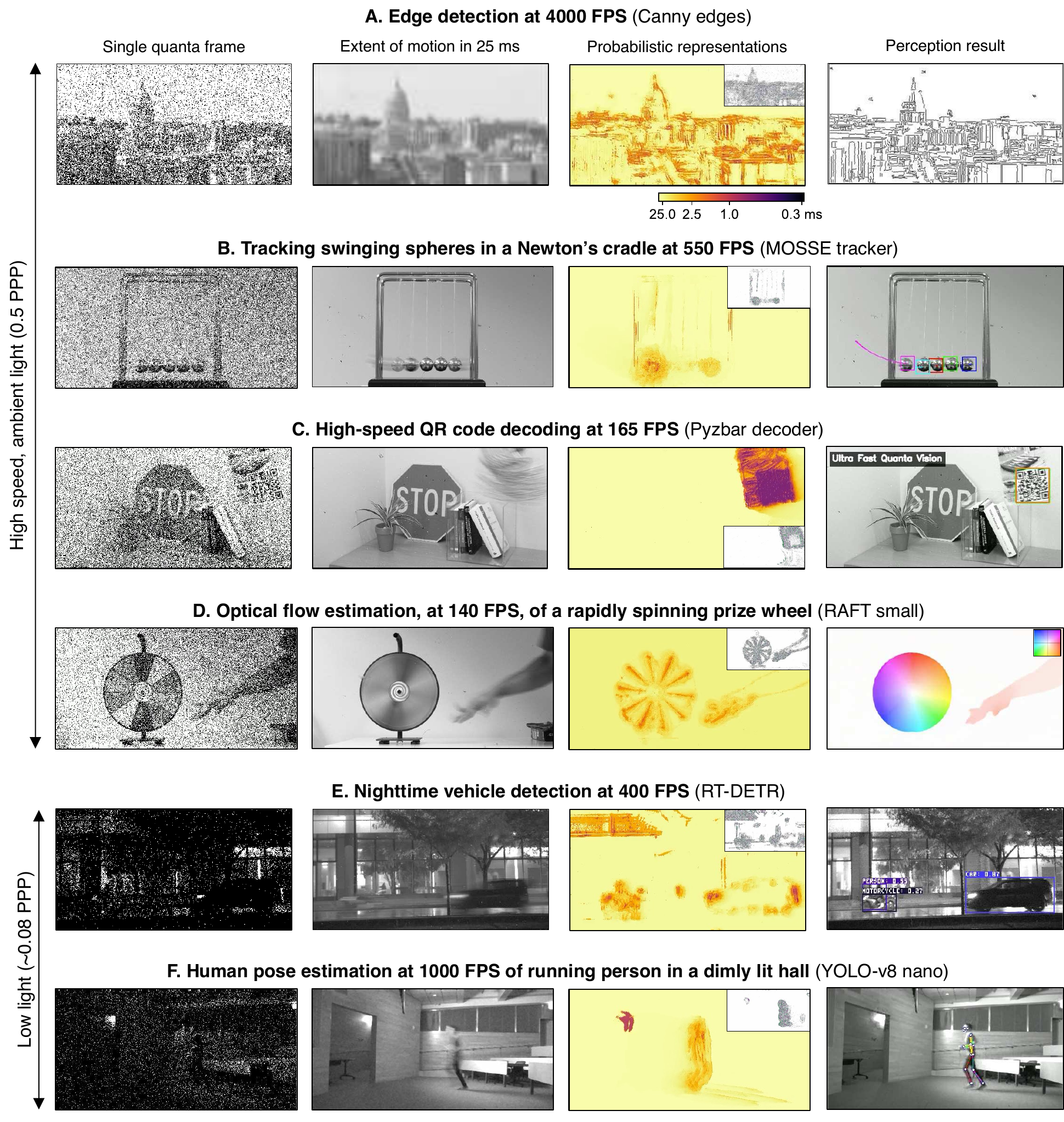}
    \vspace{-0.2in}
    \caption{\textbf{Low-latency quanta perception in the wild.} Motion-aware flux aggregates produced by probabilistic events serve as \emph{plug-and-play} inputs to standard vision pipelines, enabling robust perception in regimes where conventional cameras are constrained by the noise--blur trade-off. We evaluate tasks spanning classical vision and modern deep networks \emph{without} retraining or task-specific tuning. \textbf{(A--D) High-speed regimes:} edge detection at 4{,}000~FPS (Canny~\citep{CannyEdge1986}); tracking the swinging spheres of a Newton's cradle at 550~FPS (MOSSE~\citep{bolme2010visual}); QR-code decoding at 165~FPS;
    and optical flow on a rapidly spinning prize wheel at 140~FPS (RAFT~\citep{teed2020raft}). \textbf{(E--F) Low-light regimes:} nighttime vehicle detection at 400~FPS (RT-DETR~\citep{zhao2024detrs}); and human pose estimation of a running person in a dim hall at 1{,}000~FPS (YOLOv8~\citep{Jocher_Ultralytics_YOLO_2023}). \textbf{FPS numbers indicate end-to-end throughput}, including the time needed to compute the probabilistic event representation and run subsequent inference. Each row shows a representative \textit{single quanta frame}, the \textit{extent of motion} under a conventional 25\,ms integration, the \textit{probabilistic representation} (color indicates the inferred temporal stability / adaptive integration timescale), and the resulting \textit{perception output}. Throughputs measured while processing $256 \times 512$ quanta camera streams (edge detection and human pose estimation utilize the faster probabilistic event variant, capable of output throughputs up to 9,000~FPS).}
    \label{fig:plug_and_play}
\end{figure*}

%% file: figures/qbp_comparison.tex
\begin{figure}[t]
    \centering
    \includegraphics[width=\linewidth]{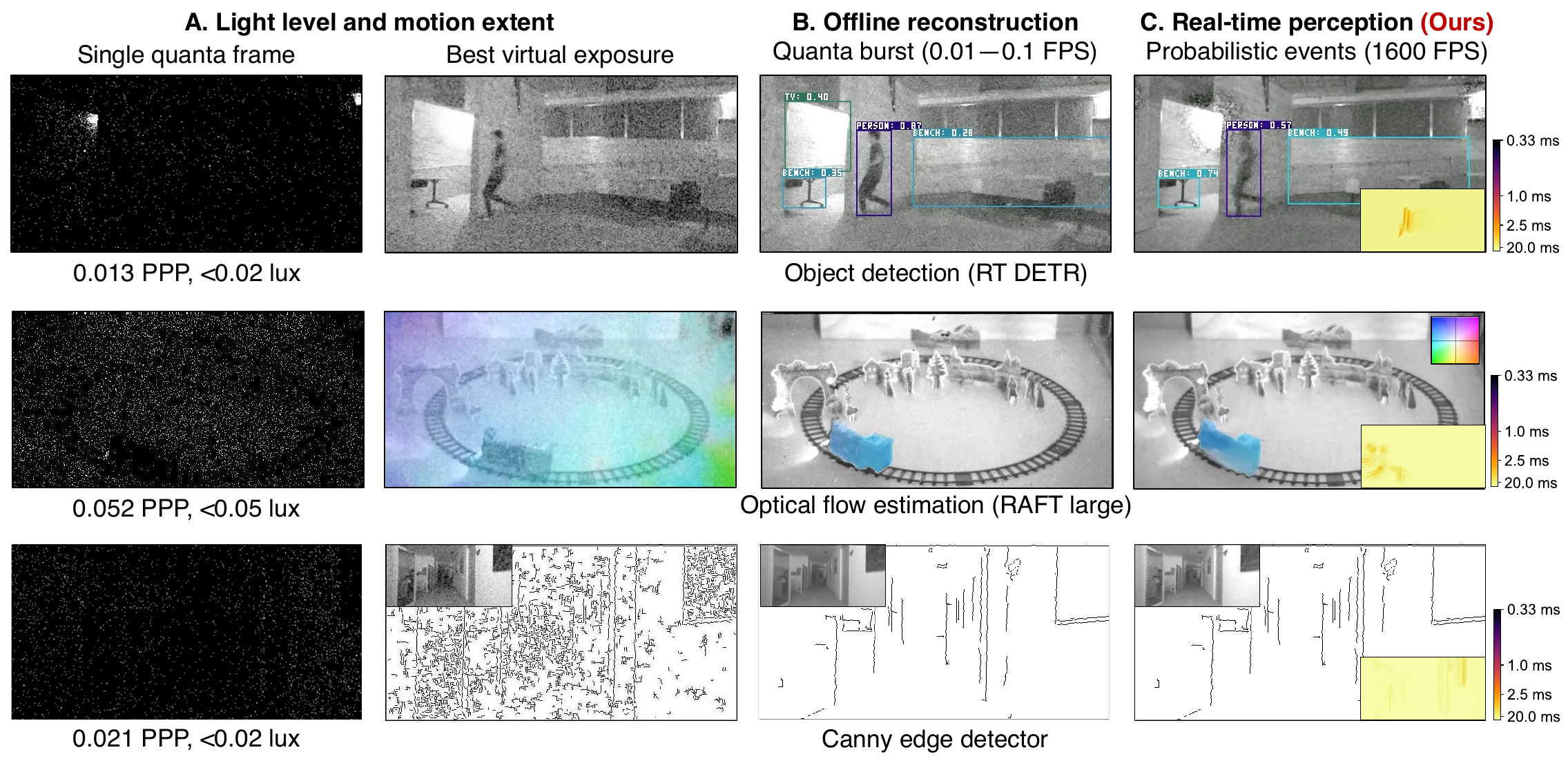}
    \vspace{-0.2in}
    \caption{\textbf{Comparison with offline quanta-reconstruction on low-light scenes.} \textbf{(A)} We consider low-light scenes in which the detector (SwissSPAD2 sensor~\protect\citep{ulku512512spad2019}) registers, on average, less than $0.05$ photons per pixel (PPP)---corresponding to less than $0.05$ lux illumination.
    Summing photon detections incurs the noise--blur trade-off at the frame level; even when this trade-off is balanced carefully, we find that virtual exposures (summed frames) do not facilitate computer-vision tasks. \textbf{(B)} An offline quanta reconstruction technique (Quanta Burst Photography~\protect\citep{ma_quanta_2020}) reconstructs high quality images by performing non-causal align-and-merge operations on the raw binary sequence, leading to good task performance, but with very high latencies. \textbf{(C)} In comparison, the probabilistic event camera balances the noise--blur trade-off at the pixel level, running significantly faster (about 10,000$\times$ faster); insets visualize the temporal stability maps. Annotated frame rates indicate the output representation rate supplied to downstream perception.}
\label{fig:qbp_comparison}
\end{figure}

%% file: figures/comparison_to_conventional_cameras.tex
\begin{figure}[t]
    \centering
    \includegraphics[width=\columnwidth]{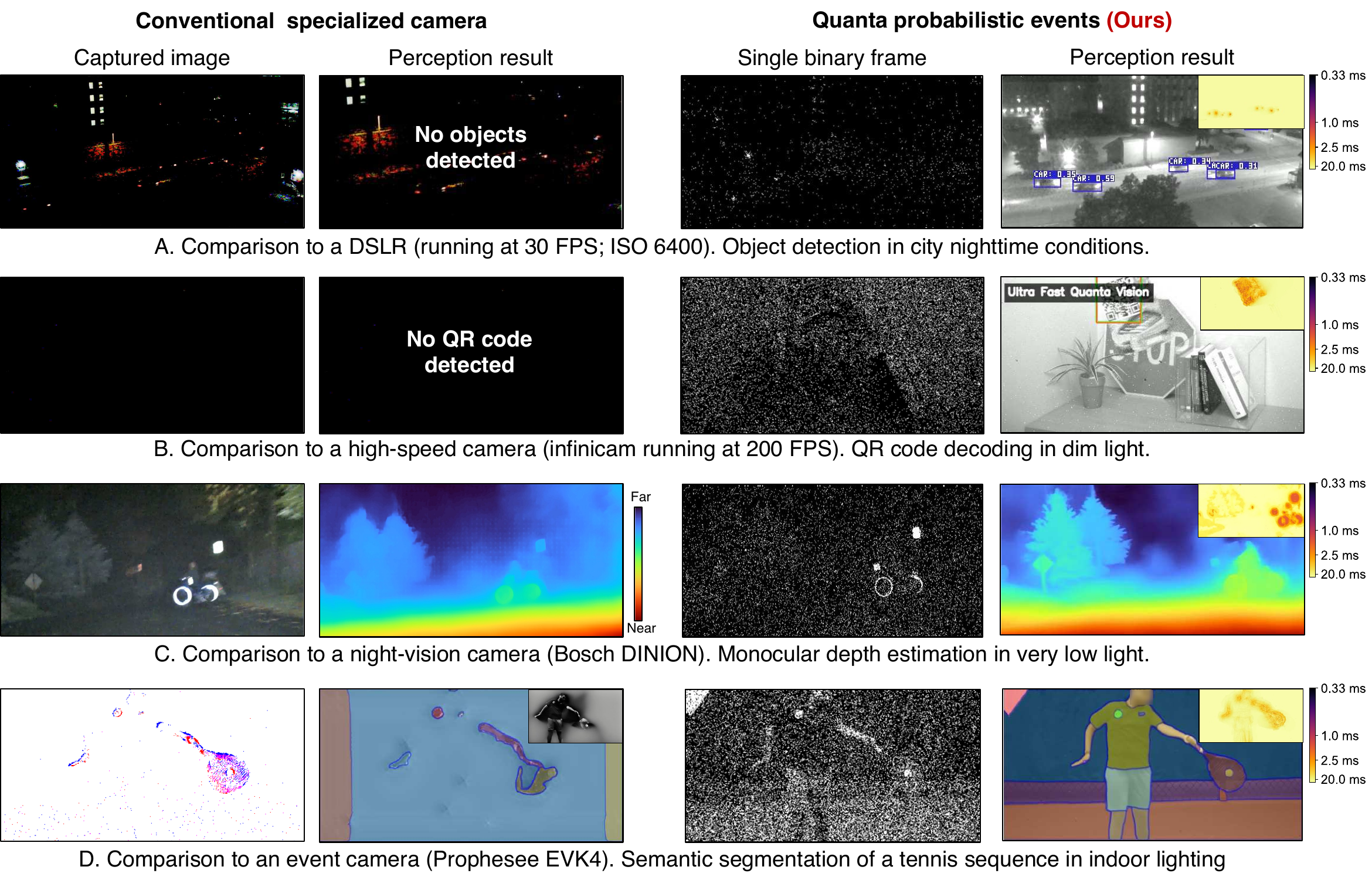}
    \vspace{-0.2in}
    \caption{\textbf{Probabilistic events compared to specialized sensing modalities.} We benchmark our computational approach applied to a quanta sensor against a conventional DSLR (Canon 77D, 30 FPS), high-speed camera (Photron Infinicam, 200 FPS), low-light camera (Bosch DINION), and event camera (Prophesee EVK4).
    We tuned conventional sensor and downstream inference parameters for near-optimal performance.
    Each row shows conventional camera capture (showing failure), a single quanta frame, and our probabilistic event representation; insets show temporal stability maps.
    \textbf{(A) City Nighttime:} The DSLR capture is insufficient for object detection (RT-DETR~\citep{zhao2024detrs}). Our method enables car detection at 100 Hz.
    \textbf{(B) High-Speed QR Decoding in Low Light:} The Infinicam fails to capture usable data at 200 FPS. Its compression quantizes the low signal-to-noise ratio (SNR) signal to zero, producing dark frames and decoding failure. In contrast, our method decodes the QR code in 6 ms.
    \textbf{(C) Low-Light Depth:} The security camera suffers motion blur from ego-motion. Probabilistic events retain structural edges (motion-aware aggregate in inset), enabling depth estimation at 100 Hz (DepthAnythingV2-s~\citep{yang2024depth}).
    \textbf{(D) Segment Anything:} Event cameras lack intensity in static regions, causing segmentation failure (SAM~\citep{Kirillov_2023_ICCV}) despite E2VID reconstruction (inset). Our approach captures full context, enabling accurate segmentation. \textit{Faces blurred for privacy.}}
\label{fig:comparison_to_conventional_cameras}
\end{figure}

%% file: sections/discussion.tex
\section*{Discussion}

\input{figures/limitations}

\subsection*{Operating Limits at the Photon Limit}

A fundamental limitation of probabilistic events, and indeed any statistical change detector, is the inability to reliably distinguish true scene dynamics from the stochasticity of photon arrivals. When the photons are too sparse, the run-length posterior can remain overly confident in a stale hypothesis, causing missed change points and over-integration. Conversely, when photon detections become near-deterministic (\eg, under very high flux or saturation effects), small model mismatches can lead to overly frequent change declarations and under-integration. In both cases, the estimator’s ability to separate motion from shot noise degrades.

Empirically, we find that the proposed estimator is most reliable across the middle 97\% of photon detection probabilities per quanta frame. Outside this regime---specifically when extreme low light coincides with very rapid motion---the effective signal-to-noise ratio drops below what is needed to confidently update the run-length posterior, and motion is increasingly absorbed into the integration window, leading to blur. \Cref{fig:limitations} illustrates a failure case in this information-limited regime.

These failure modes reflect limits of the available information in the photon streams: when the measurements do not contain enough information to disambiguate change from noise, a causal estimator must trade temporal responsiveness against noise suppression: balancing the latency of adapting to new dynamics against the variance of overreacting to stochastic noise. Conventional intensity cameras typically commit to a global exposure over each frame, enforcing a scene-wide compromise between shot noise and motion blur. In contrast, probabilistic events adapt the integration timescale locally through per-pixel Bayesian updates, allowing static regions to accumulate evidence while contracting integration around dynamic structure. As shown in \cref{fig:comparison_to_conventional_cameras} (and quantified in Methods), although subject to fundamental information-theoretic limits, transforming photon streams into actionable, high-fidelity representations enables a single quanta sensor to compare favorably against an array of specialized cameras across diverse conditions.

\subsection*{Toward Near-Sensor Probabilistic Processing}

While we demonstrate computational viability on desktop-grade systems (NVIDIA 4090 GPU), the structure of probabilistic events naturally lends itself to near-sensor deployment. Representing a step toward sensor-proximal deployment, on an embedded platform (NVIDIA Jetson Orin Nano), the gradient variant sustains $\approx 3{,}200$ qFPS at 1~megapixel (\cref{fig:metering_runtime}). The core computation consists of bounded-state Bayesian updates and exponential smoothing performed independently at each pixel, with constant-time recursion and fixed memory. Furthermore, probabilistic events feature design knobs that allow system designers to smoothly trade computational complexity for speed. For instance, the binomial formulation operates on multi-bit quanta frames efficiently aggregated from raw binary inputs; this trades excess temporal resolution for computational headroom without compromising high-speed perception. System designers can also select linear feature extractors ranging from lightweight separable filters to robust multi-scale architectures. Unlike reconstruction-first pipelines, which require large temporal buffers and dense motion alignment, probabilistic events operate causally and locally, making them compatible with emerging stacked-sensor and in-pixel processing architectures~\citep{ardelean2023computational}.

Direct inference on photon streams enables the transmission of task-ready perceptual fields rather than raw binary detections, substantially reducing bandwidth and power demands.
In this sense, probabilistic events point toward a software-defined quanta camera: a sensor that captures photons at their physical limit, but emits configurable probabilistic representations tailored to downstream tasks. By aligning representation design with hardware constraints, this formulation opens a path toward integrated sensing-and-inference systems in which the dominant computational burden shifts away from interpreting photon streams and toward higher-level perception.

\subsection*{Relation to Event Cameras}
Our formulation invites comparison with standard neuromorphic event cameras~\citep{lichtsteiner200364x64,davis240}\citep{finateau2020prophesee}. Traditional event sensors asynchronously emit binary polarity changes triggered by fixed analog thresholds. In contrast, probabilistic events maintain a posterior distribution over change intervals at each pixel. This shift from thresholded triggers to probabilistic belief states yields additional, task-relevant fields derived directly from photon statistics. In particular, the temporal stability map provides a continuous analogue of event activity, encoding both the presence and the temporal scale of change rather than a binary spike. Furthermore, the change in posterior entropy captures the evolution of uncertainty, exposing regions where the belief about temporal stability is actively shifting or resolving. Crucially, the motion-aware flux aggregate---driven by the temporal stability map---provides direct access to a high-fidelity intensity estimate. This stands in contrast to traditional event cameras, which register only the polarity of change.

Together, these signals generalize change-based sensing from discrete event detection to uncertainty-aware, image-like representations that remain compatible with standard computer vision pipelines.

Architecturally, probabilistic events are stateful and frame-based, and can be read out at arbitrary cadence. Whereas conventional event cameras prioritize sparse asynchronous transmission to reduce bandwidth, our formulation decouples internal update rate from output rate. The sensor may maintain a high-frequency internal state (\eg, $\sim$100~kHz) while delivering frames on demand at a task-appropriate rate. Moreover, because outputs are image-like fields, mature video compression pipelines (\eg, hardware-accelerated codecs) can be leveraged for efficient transmission, as demonstrated in Methods. In this sense, probabilistic events preserve the change-centric philosophy of event sensing while exploiting photon-counting measurements to produce richer, uncertainty-aware perceptual signals under tight latency constraints.

%% file: figures/limitations.tex
\begin{figure}[t]
    \centering
    \includegraphics[width=\columnwidth]{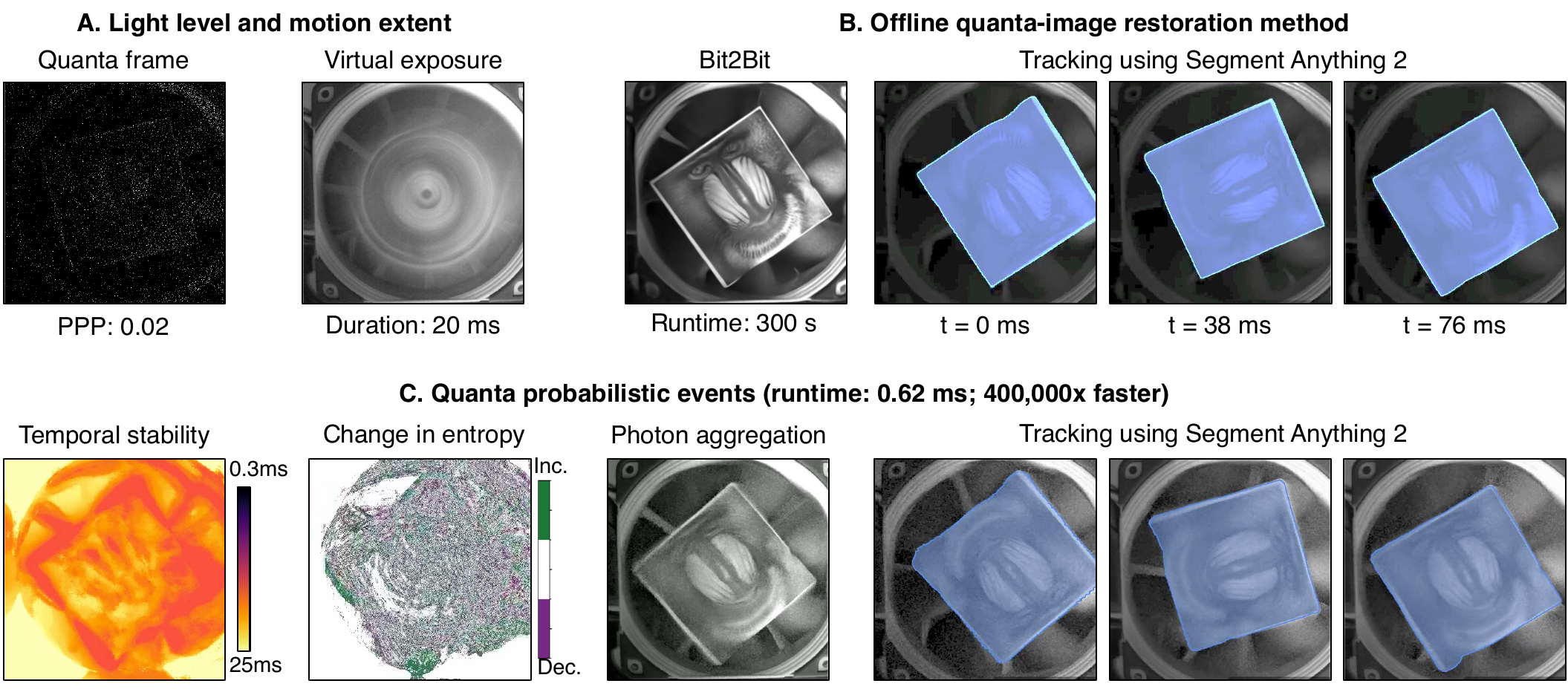}
    \vspace{-0.2in}
    \caption{\textbf{Limitations in signal-starved regimes.} \textbf{(A)} We evaluate performance in an extreme low-light scenario (0.02 photons per pixel) featuring rapid motion: a picture of a Mandrill mounted on a rotating CPU fan (sensor acquisition from \protect\citep{liu2024bit2bit}).
    \textbf{(B)} We benchmark against Bit2Bit (top)~\protect\citep{liu2024bit2bit} to define a performance ceiling. This method employs inference-time training to perform self-supervised denoising on the test data itself; while this recovers sharp details, the optimization is computationally exhaustive. We report its batch throughput as $\sim 3 \times 10^{-3}$ FPS, corresponding to roughly 300~s per output frame after optimization; because the optimization is performed over the sequence, the latency before outputs are available is a few hours.
    \textbf{(C)} In contrast, our probabilistic events operate in real-time (0.62 ms), representing a speedup of five orders of magnitude. Here, we visualize the resulting temporal stability map, entropy shifts, and flux aggregate. The trade-off for this ultra-low latency is the inability to employ iterative optimization or non-causal alignment. Consequently, our method struggles to distinguish rapid changes from shot noise, resulting in motion blur. Nevertheless, the motion-aware aggregation preserves discernible structure; notably, Segment Anything 2 (SAM 2, \cite{ravi2024sam}) successfully tracks the object despite the aggregate being an imperfect reconstruction.}
    \label{fig:limitations}
\end{figure}

%% file: sections/methods.tex
\section*{Methods}\label{sec:methods}

To overcome the fundamental limitations of fixed-exposure integration---which inherently entangles noise, blur, and dynamic range---we formulate probabilistic events, a computational pipeline that transforms the stochastic arrival of individual photons into motion-adaptive visual streams. We begin by presenting the image formation model of quanta sensors, demonstrating the fundamental trade-off between noise and motion blur. To navigate this trade-off, we introduce the core mechanism of the probabilistic event camera, detailing how Bayesian inference drives per-pixel exposure adaptation through recursive run-length estimation. Finally, we expand this primitive into a complete perception pipeline; while our core representation supports direct machine vision, we also detail its integration with spatially varying denoising, hardware-accelerated video compression, and color imaging to provide a comprehensive imaging solution that bridges the gap between photon-counting physics and practical applications.

\subsection*{Image Formation in Quanta Cameras}\label{sec:background}

Quanta cameras sense light as streams of photon detections, specifically, binary-valued frames captured at high speeds ($\approx 100$~kHz) synchronized to an internal clock. Each binary value indicates whether at least one photon was detected during that specific exposure interval. If $\phi_t(\vp)$ denotes the average photon count incident on the sensor at pixel $\vp$ and frame $t$, these binary responses, $B_t(\vp)$, can be modeled as Bernoulli random variables (r.v.s)~\citepmethods{yang_poisson_model}:
\begin{equation}\label{eq:spad_model}
    \Prob{\curly{B_t(\vp) = 1}} = 1 - e^{-\phi_t(\vp)}.
\end{equation}
\Cref{fig:noise_blur}A shows an example of a binary quanta frame. While our exposition focuses on this binary photon-detection mode, the framework generalizes to quanta sensors that capture multiple photons per exposure, such as those based on Jots~\citepmethods{fossum:11,fossum2016quanta,ma:17,ma2022review}.

To estimate scene intensity, we sum binary-valued frames across time, creating a \textit{virtual exposure} that mimics how conventional sensors gather photons:
\begin{equation}\label{eq:sum_image}
    S_t(\vp) = \sum_{\tau= t -N_\text{sum}}^t B_\tau(\vp),
\end{equation}
where $N_\text{sum}$ represents the exposure time. When $\phi_t(\vp)$ is constant over this duration, implying negligible scene or camera motion, $S_t(\vp)$ is the sum of independent and identically distributed (i.i.d.) Bernoulli variables. Consequently, it follows a binomial distribution, enabling maximum likelihood estimation of scene flux~\citepmethods{antolovic:18}.

Crucially, quanta sensors allow us to select the exposure time, $N_\text{sum}$, \textit{post hoc}~\citepmethods{ma_2023_wacv,Jungerman_2023_ICCV}. However, this choice is strictly governed by scene dynamics, forcing a fundamental compromise. As demonstrated in \cref{fig:noise_blur}B and \cref{fig:noise_blur}C, a globally fixed exposure time fails to adapt to these dynamics: summing too few frames (N = 32) yields noisy images, while summing too many (N = 8000) introduces severe motion blur.

To mitigate this trade-off, quanta-image reconstruction techniques often apply motion compensation before merging frames~\citepmethods{ma_quanta_2020,iwabuchi2021,chennuri2025quanta}. Unfortunately, the prohibitive computational cost of explicit motion compensation precludes real-time reconstruction, limiting the utility of quanta sensors in latency-critical applications.

Can we omit motion compensation in the pursuit of speed? In the next section, we describe a probabilistic formulation where each pixel automatically determines its temporal stability based on expected motion. This maximizes temporal denoising without explicit motion alignment, achieving a favorable noise--blur balance with significant computational advantages.

\subsection*{Computing Temporal Stability Maps}\label{sec:choosing_exposures}

We now describe the core mechanism of the \textit{probabilistic event camera}. Unlike standard event cameras that strictly report binary brightness changes, our approach models the probability distribution of signal stability to drive a recursive photon aggregator. We begin with the base formulation operating on raw Bernoulli detections (as used in prior works~\citepmethods{Sundar_2024_CVPR,Sundar_2025_ICCV}) before devising more computationally efficient generalizations.

Our estimator computes an exponential moving average (EMA) of photon detections, a streaming variant of \cref{eq:sum_image}:
\begin{equation}\label{eq:ema}
    \mathcal{I}_{\text{EMA}, t}(\vp) = (1 - \omega) \,\mathcal{I}_{\text{EMA}, t - 1}(\vp) + \omega B_t(\vp),
\end{equation}
where $\omega \in [0, 1]$ acts as a sample weight that dictates temporal stability. Smaller values of $\omega$ correspond to longer periods of assumed stability, resulting in stronger smoothing: specifically \begin{equation}\label{eq:window_size_to_sample_weight}
    \omega = \frac{2}{N + 1}
\end{equation}
yields a steady-state output noise variance equivalent to that of a uniform box filter of length $N$~\citepmethods{brown2004smoothing}. The advantage of exponential smoothing is its recursive nature, which allows us to compute a video of integrated frames without redoing computations for each frame. Furthermore, the inherent linearity of this exponential smoothing naturally accommodates the expectation operations we later introduce in \cref{eq:adaptive_ema_derived}.

To achieve spatially and temporally adaptive smoothing \citepmethods{trigg1967exponential}, we let $\omega$ vary with time $t$ and pixel location $\vp$:
\begin{equation}\label{eq:adaptive_ema}
    \mathcal{I}_{\text{adapt}, t}(\vp) = (1 - \omega_t(\vp)) \,\mathcal{I}_{\text{adapt}, t - 1}(\vp) + \omega_t(\vp) B_t(\vp).
\end{equation}
The challenge lies in actively estimating the temporal stability of each pixel to dynamically set $\omega_t(\vp)$ based on motion. This estimation task requires causal prediction: we must efficiently model incident detections without storing long histories.

To this end, we employ Bayesian Online Change Point Detection (BOCPD)~\citepmethods{adams2007bayesian} to estimate the time since the last abrupt change, or the \textit{run length}. At each pixel, the algorithm calculates the posterior probability distribution $\Prob{r_t \mid B_{1:t}(\vp)}$ over the run length $r_t$. BOCPD updates this distribution via recursive message passing. Omitting the spatial index $\vp$ for brevity, the probability that the run length grows is
\begin{equation}\label{eq:message_passing_growth}
    \Prob{r_t = r+1 \mid B_{1:t}} \propto  (1-\gamma) \Prob{B_t \mid r_{t-1}=r, B_{1:t-1}} \Prob{r_{t-1} = r \mid B_{1:t-1}},
\end{equation}
while the probability of a change point (run length resets to zero) is
\begin{equation}\label{eq:message_passing_change}
    \Prob{r_t = 0 \mid B_{1:t}} \propto \sum_{r=0}^{t-1} \gamma \Prob{B_t \mid r_{t-1}=r, B_{1:t-1}}  \Prob{r_{t-1} = r \mid B_{1:t-1}},
\end{equation}
where $\gamma \in [0, 1]$ is the hazard rate (we set $\gamma=10^{-5}$). \cref{eq:message_passing_growth,eq:message_passing_change} imply that run lengths either increment by one or reset. This assumption induces a trellis structure on the inference process, a constraint essential for online efficiency~\citepmethods{adams2007bayesian}.

\subsubsection*{Bernoulli Instantiation}
We use a Bernoulli likelihood with a conjugate Beta prior to enable exact, closed-form recursive updates. The predictive probability for a segment having run length $r$ at time $t-1$ depends on hyperparameters initialized at the segment's start, $s = t-1-r$:
\begin{equation}\label{eq:posterior_predictive_bernoulli}
    \Prob{B_t = 1 \mid r_{t-1} = r, B_{1:t-1}} = \frac{\alpha_s}{\alpha_s + \beta_s}.
\end{equation}
New segments ($r_t=0$) are initialized with a non-informative Jeffrey's prior, $\alpha_t = \beta_t = 0.5$. Upon observing $B_t$, we update the parameters for each potential start time $s < t$:
\begin{equation}\label{eq:param_update_bernoulli}
    \alpha_s \gets \alpha_s + B_t,\quad
    \beta_s \gets \beta_s + 1 - B_t.
\end{equation}

\subsubsection*{Bounding Memory Usage}
Naïve BOCPD implies linear memory growth, storing probabilities and hyperparameters for all $t$ potential run lengths. To prevent this linear growth, we employ stratified pruning, storing only the top-$K$, typically $5$--$8$, run-length probabilities at each step~\citepmethods{wang2021online}. More precisely, if $\mathcal{P}_\text{sorted}$ denotes the sequence of run-length probabilities, $\Prob{r_t \mid B_{1:t}} \;\forall\; r_t \in \curly{0, 1, ..., t-1}$, sorted in descending order, we retain and renormalize the first $K$ entries. This approximation is effective because run-length probabilities naturally concentrate around a few likely values.

\subsubsection*{Estimating the Expected Adaptive Decay}
Given the run-length distribution, we must derive an estimator for the temporal stability parameter $\omega_t(\vp)$. A direct approach might map run-length directly to a \textit{stochastic stability metric} $\Omega_t(\vp) = 2/(r_t(\vp)+2)$, but this would require maintaining a distribution of aggregation hypotheses, that is, an aggregate $\mathcal{J}_t(\vp)$ corresponding to each $\Omega_t(\vp)$ (and by extension, $r_t(\vp)$). Instead, we compute the \textit{expected} aggregate, $\mathcal{I}_{\text{adapt}, t}(\vp) \equiv \mathbb{E}[\mathcal{J}_t(\vp)]$.

By taking the expectation of the recurrence and leveraging the independence of the new weight $\Omega_t(\vp)$ from the past aggregate $\mathcal{J}_{t - 1}(\vp)$ (conditioned on history), we derive:
\begin{equation}\label{eq:adaptive_ema_derived}
    \mathcal{I}_{\text{adapt}, t}(\vp) = (1 - \omega_t(\vp)) \,\mathcal{I}_{\text{adapt}, t - 1}(\vp) + \omega_t(\vp) B_t(\vp),
\end{equation}
where $\omega_t(\vp) \equiv \mathbb{E}[\Omega_t(\vp)]$. Practically, this derivation means we estimate $\omega_t(\vp)$ as the expected temporal stability over the computed run-length distribution. We use the updated weighting scheme ($2/(n+1)$ where $n=r+1$) to minimize ghosting artifacts:

\begin{equation}\label{eq:omega_estimation}
    \omega_t(\vp) = \sum_{r=0}^{t} \frac{2}{r+2} \Prob{r_t = r \mid B_{1:t}(\vp)}.
\end{equation}
This expectation-based formulation yields a temporally denoised output that aggregates detections based on continuous stability. Crucially, while the underlying inference strictly restricts individual run-length hypotheses to discrete state transitions (incrementing by one or resetting to zero), taking the expectation over this posterior distribution yields a continuous metric. This mathematical property allows our formulation to model smooth intensity transitions---such as soft gradients or continuous lighting changes---rather than being constrained to abrupt, step-like shifts. Unlike standard fixed-exposure summation, the probabilistic event camera dynamically optimizes the noise--blur trade-off per pixel. By constructing temporal stability maps from the estimated $\omega_t(\vp)$ (\cref{fig:noise_blur}D), we observe this adaptation in practice: the model assigns short exposures (~2.5 ms) to the spinning prize wheel to freeze motion, while accumulating light for 20 ms or more in the static background (\cref{fig:noise_blur}F).

\subsubsection*{Quantifying Uncertainty via Run-Length Entropy}

We quantify the estimator's uncertainty via the Shannon entropy of the run-length distribution:
\begin{equation}\label{eq:entropy}
H_t(\vp) = - \sum_{r=0}^t \Prob{r_t = r \mid B_{1:t}(\vp)} \log \Prob{r_t = r \mid B_{1:t}(\vp)}.
\end{equation}
While the raw entropy level indicates absolute uncertainty, we find that \textit{temporal differences} of entropy provide a more salient signal for scene dynamics. As an edge traverses a pixel, the local complexity spikes, causing a sharp increase in uncertainty; this is typically followed by a decrease as the estimator stabilizes on the subsequent smooth region. Consequently, significant changes in entropy serve as a robust detector for moving edges and corners, characterized by a distinct signature: a leading front of rising uncertainty followed by a wake of stabilization. As seen in \cref{fig:noise_blur}E, this differential entropy map highlights complex motion boundaries and regions of low confidence, providing a valuable secondary signal alongside the intensity estimate.

\subsubsection*{Extension to Binomial Frames}

To obtain the computational headroom necessary for more sophisticated modeling approaches, such as the spatial features introduced in the next section, we operate on \textit{virtual exposures}, $S_t(\vp)$. These are created by summing $N_\text{sum}$ consecutive, non-overlapping binary frames.

Summing binary frames is relatively inexpensive and boosts input throughput. For instance, while the Bernoulli instantiation runs at $\approx 10,000$ qFPS, the binomial instantiation (using $4$-bit frames from summing 15 binary frames) supports inputs of $\approx 150,000$ qFPS. This comes at the cost of potentially introducing minor motion blur since the minimum exposure time increases; however, for many practical computer vision applications, this minimum exposure time (approximately $0.1$--$0.3$ ms) is still considered ultra-fast.

We re-index time such that $t$ refers to these summed frames and apply BOCPD to the sequence $\{S_1, S_2, \dots\}$. For binomial data, the predictive distribution follows a Beta-binomial form. The probability of observing $S_t = n$ given a run of length $r$ is:
\begin{equation}\label{eq:posterior_predictive_binom}
    \Prob{S_t = n \;\mid\; r_{t-1}=r, S_{1:t-1}} = \binom{N_\text{sum}}{n} \frac{B(\alpha_s + n, \beta_s + N_\text{sum} - n)}{B(\alpha_s, \beta_s)}.
\end{equation}
The hyperparameters are updated as:
\begin{equation}\label{eq:param_update_binom}
        \alpha_s \gets \alpha_s + S_t, \quad
        \beta_s \gets \beta_s + N_\text{sum} - S_t.
\end{equation}

\subsection*{Incorporating Spatial Information}

The purely temporal nature of the Bernoulli and binomial probabilistic event models has its limitations. Confined to individual pixel histories, the model must accumulate sufficient local evidence before the posterior distribution can confidently shift, inherently delaying adaptation. For example, within the textureless interiors of moving objects, a pixel observes uniform photon arrivals until a structural edge physically traverses it; responding only after this traversal introduces local motion blur. Ultimately, purely temporal inference is bounded by the speed of point-wise evidence accumulation.

We therefore generalize the probabilistic event camera to model linear spatial features, such as gradients and filter-bank responses (\eg, Log-Gabor). By pooling neighborhood context, these features allow pixels to anticipate and respond to abrupt transitions even before the moving boundary directly crosses them. 
We incorporate spatial information by applying linear filters to the virtual exposures $S_t(\vp)$ defined previously, producing a feature tensor $F_t$ where each pixel $\vp$ has a vector $F_t(\vp, \cdot) \in \rr^{\texttt{C}}$.

We variance-stabilize $S_t(\vp)$ using an arcsine transform~\citepmethods{yu2009variance} and model the feature vector as a multivariate Gaussian, $F_t(\vp, \cdot) \sim \mathcal{N}(\boldsymbol{\mu}_s, \sigma_s^2 \Sigma_s)$. For computational reasons, we handle the covariance matrix $\Sigma_s$ (which captures filter correlations) via offline eigendecomposition: $\Sigma_s = \boldsymbol{V} \boldsymbol{\Lambda} \boldsymbol{V}^\top$. We project raw features onto the orthonormal basis $\boldsymbol{V}$, yielding decorrelated features $\tilde{F}_t = \boldsymbol{V}^\top F_t$. This projection diagonalizes the covariance ($\Sigma_s \to \boldsymbol{\Lambda}$), allowing independent Gaussian updates per channel. This transformation also facilitates compression, enabling us to retain only leading principal components when using high-dimensional filter banks.

Assuming independence between summed intensity and spatial features given the run length, the joint log-predictive distribution is:
\begin{equation}\label{eq:posterior_predictive_joint}
\log \Prob{\tilde{F}_t = \boldsymbol{f}, S_t = n \mid \dots} = \log \Prob{\tilde{F}_t = \boldsymbol{f} \mid \dots} + \log \Prob{S_t = n \mid \dots}.
\end{equation}
The feature likelihood follows the posterior predictive of the multivariate Gaussian:
\begin{equation}\label{eq:posterior_predictive_gaussian}
\Prob{\tilde{F}_t = \boldsymbol{f} \mid r_{t-1}=r, \dots} \propto \mathcal{N}\left(\boldsymbol{f}; \boldsymbol{\mu}_{s}, \sigma_s^2 \boldsymbol{\Lambda}\right),
\end{equation}
with $\sigma_s^2 = 1 + 1/\paren*{r+1}$. The mean $\boldsymbol{\mu}_{s}$ updates recursively:
\begin{equation}\label{eq:param_update_gaussian}
    \boldsymbol{\mu}_{s}(\vp) \gets \boldsymbol{\mu}_{s}(\vp) + \frac{\tilde{F}_t(\vp, \cdot) - \boldsymbol{\mu}_{s}(\vp)}{r+1}.
\end{equation}

\subsubsection*{Choosing Linear Features}

We evaluate linear feature extractors to balance computational cost and spatial support. While $5$-tap separable gradients~\citepmethods{Farid:2004:Differentiation} offer the lowest computational footprint, multi-scale filter banks like Log-Gabor and Difference of Gaussians (DoG) robustly capture spatial structures. We implement a separable DoG bank acting as a bandpass filter~\citepmethods{kovesi2010fast}, parameterizing it to approximate the spectral coverage of a 4-scale Log-Gabor bank with a minimum wavelength of $3.0$ pixels. Unlike the explicitly oriented Log-Gabor bank, our DoG formulation projects directional changes onto separable vertical and horizontal components. This yields a representation that is more compact and computationally efficient, albeit with reduced directional sensitivity.

\Cref{fig:using_spatial_info} highlights the inverse relationship between responsivity and spatial precision. The space-agnostic binomial model often exhibits sluggish responses to changes, resulting in motion blur, though it remains spatially precise. A simple heuristic to increase responsivity is to artificially truncate the temporal stability map, for example, via minimum filtering, as a post-hoc measure~\citepmethods{Sundar_2025_ICCV}. However, this approach is content-blind; it forces shorter exposures even in slow-moving regions, significantly increasing noise in the flux aggregate. Using gradients in the run-length modeling better incorporates spatial content, producing sharper temporal stability maps. The DoG formulation adds robustness via scale information, reducing noise compared to gradients, but is susceptible to ringing artifacts near edges. Finally, Log-Gabor filters are highly sensitive to structure but often suffer from blooming---a characteristic artifact of frequency-domain filters---where responsivity dilates motion boundaries and increases noise in nearby static regions.

Ultimately, we err on the side of responsivity: we accept the slight noise penalty around moving objects to ensure motion is frozen, as residual noise is more easily mitigated by downstream denoisers than deleterious motion blur. Modeling spatial features selectively increases pixel responsivity in moving regions. In this example, our techniques offer input and output throughputs ranging between 51,000--280,000 qFPS and 1,600--9,000 FPS, respectively, when operating on 5-bit virtual exposures ($N_\text{sum}=31$). Moving forward, we focus on two representative configurations: we designate the gradient-based formulation as our \textit{fast} variant, and the Log-Gabor formulation as our \textit{robust} variant.

\subsubsection*{Runtime Analysis}

By aggregating $N_\text{sum}=31$ binary frames---effectively treating a high-bandwidth $100$~kHz binary sensor as an $\approx 3$~kHz multi-bit camera---we amortize the overhead of spatial filtering.
As shown in \cref{fig:metering_runtime}, our implementation exploits massive parallelism. Notably, the Gradient, Binomial, and Bernoulli variants exhibit a runtime plateau, maintaining effectively constant throughput between $256 \times 256$ and $512 \times 512$ resolutions. At $1024 \times 1024$, runtime increases but scales sublinearly, indicating that performance is overhead-bound at lower resolutions and only begins to saturate memory bandwidth at the megapixel scale. All results were obtained using \texttt{torch.compile} on an NVIDIA 4090 GPU; however, we anticipate that further gains could be realized through custom CUDA kernels that fuse operations and persist state in registers, minimizing costly transfers to global memory.

Efficiency is most pronounced in our \textit{fast} variant (binomial + gradients), which achieves input and output throughputs of $\approx 41,000$ qFPS and $1,280$ FPS, respectively, for a 1 MPixel sensor when using 5-bit virtual exposure inputs. Conversely, the \textit{robust} variants incur a higher cost due to multi-scale filtering yet remain viable for real-time applications. The Difference of Gaussians (DoG) filter sustains input throughputs of $\approx 260,000$ qFPS at $256 \times 256$, decreasing to $\approx 16,000$ qFPS at 1 MPixel. The more computationally intensive Log-Gabor filter follows a similar trend, scaling from $\approx 115,000$ qFPS at $256 \times 256$ to $> 6,000$ qFPS at 1 MPixel. This demonstrates a flexible system design, allowing users to modulate the trade-off between structural fidelity and raw speed according to application requirements.

We further evaluate on a power-constrained embedded platform (NVIDIA Jetson Orin Nano) to assess edge device viability (\cref{fig:metering_runtime}B). On memory-bandwidth-limited hardware such as the Jetson, lighter formulations are preferable; the gradient-based variant sustains $> 10{,}000$ qFPS up to $512 \times 512$ resolution and $\approx 3{,}200$ qFPS at 1 MPixel, offering a practical operating point for edge deployment.

\subsection*{Quanta Perception Pipeline}

We explore integrating a probabilistic event camera into the processing pipeline of a quanta sensor. In this role, it functions as a low-latency atomic primitive that enables direct visual perception, supports optional denoising for human visualization, and facilitates bandwidth-efficient data transfer.

\subsubsection*{Direct Perception}

Across a broad spectrum of light and motion levels, the motion-aware photon aggregations produced by probabilistic events possess sufficient fidelity for direct consumption by downstream computer vision algorithms, requiring neither retraining nor fine-tuning. While performance could potentially be enhanced by fine-tuning downstream models or adopting specialized neural architectures for quanta cameras~\citepmethods{Sundar_2025_ICCV}, this path presents practical challenges. Obtaining ground-truth annotations at the extreme frame rates native to quanta sensors is not trivial. Acquiring object bounding boxes at these speeds demands exhaustive manual annotation for real data or the use of specialized simulators for synthetic data; consequently, this often necessitates relying on pseudo-ground truth from teacher models running on high-speed video. To bypass these engineering bottlenecks and ensure broad compatibility with the existing ecosystem, we opt for a plug-and-play approach that grants immediate access to a wide range of pre-trained models. We validate this utility across a suite of tasks, including object detection, pose estimation, optical flow, and tracking.

\subsubsection*{Spatially-Varying Denoising}
A direct consequence of motion-adaptive integration is spatially heterogeneous noise: dynamic regions, having integrated fewer photons, exhibit higher variance than their static counterparts. To mitigate this without compromising temporal resolution, we employ a spatially-varying Wiener filter directly modulated by the temporal stability map, $\omega_t(\vp)$.

We first approximate the per-pixel noise variance $\sigma^2_{\text{noise}, t}(\vp)$ by deriving the effective run length $\tilde{r}_t(\vp) = \frac{2}{\omega_t(\vp)} - 1$. This allows us to model the noise magnitude as:
\begin{equation}
    \sigma^2_{\text{noise}, t}(\vp) = N_\text{sum} \tilde{r}_t(\vp) \cdot \mathcal{I}_{\text{adapt},t}(\vp) \cdot \paren*{1-\mathcal{I}_{\text{adapt},t}(\vp)}.
\end{equation}
Subsequent filtering relies on local statistics---mean $\mu_{\text{local}, t}$ and variance $\sigma^2_{\text{local}, t}$---computed via a Gaussian kernel. We determine the signal preservation ratio:
\begin{equation}
    R_{\text{var}, t}(\vp) = \max\paren*{0, 1 - \frac{\sigma^2_{\text{noise}, t}(\vp)}{\sigma^2_{\text{local}, t}(\vp)}},
\end{equation}
which yields the final denoised output by suppressing fluctuations that fall below the estimated noise floor:
\begin{equation}\label{eq:wiener_denoising}
    \mathcal{I}_{\text{denoised}, t} = \mu_{\text{local}, t} + R_{\text{var}, t} \cdot \paren*{\mathcal{I}_{\text{adapt}, t} -\mu_{\text{local}, t}}.
\end{equation}
As visualized in \cref{fig:spatially_varying_denoising}, this formulation aligns with intuition: regions characterized by short integration windows undergo more aggressive spatial smoothing. Crucially, this filter operates at $\sim$4,000~FPS, ensuring it does not become a system bottleneck. While alternative restoration paradigms such as neural post-processing~\citepmethods{zhang2024streaming,Sundar_2024_CVPR,chennuri2025quanta}, burst fusion, or optical-flow-guided align-and-merge~\citepmethods{seets_2021_wacv} might offer higher fidelity, they typically fail to sustain the kilohertz-range throughput inherent to our probabilistic event representations, thereby undermining the system's low-latency objectives.

\subsubsection*{Hardware-Accelerated Video Encoding}

By outputting low-latency image representations rather than asynchronous event streams, probabilistic event cameras are inherently compatible with hardware-accelerated video codecs. We empirically verify that the GPU-based NVENC H.264 encoder can process $256 \times 512$ video streams at approximately 6,000~FPS, which is capable of accommodating the output throughputs of our probabilistic events.

To evaluate this approach, we utilized a photon cube simulated from a high-speed video, processed via probabilistic events: outputting motion-aware flux aggregates that are subsequently denoised using our proposed spatially-varying Wiener filter. As illustrated in \cref{fig:rate_distortion_events_vs_video_codec}, we compare two transmission strategies: a delta-based event transmission scheme and a hardware-accelerated video codec. The event-based approach transmits an initial dense frame followed by sparse inter-frame differences, encoded using Coordinate Format (COO) with pixel locations, frame indices, and an 8-bit payload. Conversely, the video codec approach feeds the denoised aggregates directly into a standard encoder. We evaluate these methods by generating rate-distortion curves, defining rate as the compression factor relative to a naive binary photon stream, and measuring distortion via Peak Signal-to-Noise Ratio (PSNR) and Structural Similarity Index (SSIM). The event-based curve was obtained by linearly varying the difference threshold from 0.02 to 1.0, while the video codec curve was generated by sweeping the Constant Rate Factor (CRF) from 24 to 51.

As anticipated, the video codec achieves one to two orders of magnitude higher compression ratios than the event-based approach for comparable quality. Our objective here is not to present a definitive hierarchy, but rather to highlight an expanded design space: probabilistic events decouple the event-based sensing philosophy from bandwidth reduction, rendering sparse transmission as one of several viable strategies rather than a strict necessity. The optimal choice ultimately depends on the hardware topology. Simple event-based delta encoding remains exceptionally cheap to compute, making it well-suited for strictly on-sensor compression~\citepmethods{Sundar_2024_CVPR}. Conversely, as data moves off-sensor to edge platforms or dedicated photon processing units, standard video compression suites provide a powerful alternative. By navigating these different performance-to-compression trade-offs, we can effectively mitigate the data bandwidth bottleneck inherent to quanta sensors when broader computational budgets are available.

\subsubsection*{Color Quanta Perception} 

Integrating color perception necessitates a Color Filter Array (CFA) on the sensor, which in turn introduces the circular dependency of joint demosaicking and denoising. Standard demosaicking techniques are ill-suited to heavily quantized and noisy quanta frames, whereas spatiotemporal denoising typically cannot operate directly on mosaicked data.

Our design addresses this challenge by strictly decoupling per-pixel temporal modeling from spatial feature extraction. We execute the per-pixel components---specifically the binomial modeling---directly on the raw mosaicked quanta frames. To facilitate spatial feature extraction, we employ the lightweight Malvar-He-Cutler algorithm~\citepmethods{malvar2004high} to generate a provisional luminance estimate. While approximate, this estimate is sufficient to inform the Bayesian run-length model regarding spatial structure. After spatial features are extracted, the probabilistic event pipeline runs on a per-pixel basis, producing temporally denoised mosaic frames, which are converted to color images using standard off-the-shelf debayering. We also extend our spatially-varying Wiener filter to this color output by calculating local statistics independently for each spectral channel, while broadcasting the scalar temporal integration map to govern the noise floor. This broadcasting ensures regularization is consistent with the shared geometric confidence of the scene. \Cref{fig:color_metering} demonstrates the probabilistic event camera supporting Bayer-mosaicked acquisition.

\subsection*{Analyzing Probabilistic Events}\label{sec:analysis}

We structure our analysis to distinguish the practical efficacy of our specific algorithmic implementation from the fundamental potential of the sensing strategy. First, we evaluate the accuracy of our probabilistic events in determining temporal stability, benchmarking it against an optimized global shutter baseline. Here, we quantify performance in photographic stops, providing a tangible metric for how effectively the system balances the noise--blur trade-off compared to conventional sensors. Second, we establish the theoretical upper bound on SNR improvement. By comparing an ideal per-pixel shutter against the best possible global shutter, we define the performance ceiling for probabilistic event cameras, quantifying the maximum opportunity cost incurred by adhering to global exposure constraints.

\subsubsection*{How Well Do We Estimate Temporal Stability?}

To quantify the accuracy of our probabilistic temporal stability estimation, we benchmark temporal stability maps against ground truth windows derived from simulated photon streams. We generate these streams by interpolating 40~ms sequences from the i2k dataset~\citemethods{chennuri2025quanta} to 96~kHz using RIFE~\citemethods{huang2022real}. To establish ground truth, we apply our probabilistic event model to these clean high-speed sequences, treating intensities as Gaussian random variables with negligible variance. We benchmark performance against an ``oracle global shutter''---a theoretical baseline for conventional sensors that selects the single fixed integration time minimizing aggregate error. We report error in stops, defined as the base-2 logarithmic difference between the estimated and target window durations, distinguishing between over-exposure (motion blur) and under-exposure (noise). While ideal estimation requires zero error, practical systems must navigate an inherent trade-off; we prioritize minimizing motion blur, as it is typically more difficult to mitigate downstream than noise.

The advantage of per-pixel adaptation becomes evident when contrasting scenes with uniform pixel velocity against those exhibiting multiple distinct velocities. In scenes dominated by camera ego-motion (\cref{fig:analysis_exposure_time}A), where pixel velocities are relatively uniform, the oracle global shutter performs comparatively well. Because optimal integration windows are spatially correlated, the global baseline selects a compromise that yields a Root Mean Square Error (RMSE) of 1.01 stops for motion blur and 1.48 stops for noise. However, this compromise fails in scenes with distinct independent velocities, such as a card shuffling sequence (\cref{fig:analysis_exposure_time}B). Here, the bimodal distribution of optimal temporal stability forces the global baseline into a middle ground that blurs the moving hands while rendering the background noisy, resulting in a high blur penalty of 1.68 stops. Our estimator decouples these dynamics, assigning short integration windows strictly to the falling cards while extending integration for the static background. This local adaptation reduces blur RMSE to 0.26 stops; by eschewing rigid global constraints, the system dynamically manages the noise--blur trade-off at the pixel level for every instant.

Across the full dataset, our binomial variant recovers temporal stability maps with a sharply peaked error distribution centered near zero (\cref{fig:analysis_exposure_time}C). While pure binomial estimation can retain residual blur, alternative mitigation strategies like minimum filtering~\citemethods{Sundar_2025_ICCV} result in an overly responsive estimator with excessive noise. By instead incorporating spatial features (Log-Gabor), we achieve a discriminative balance that suppresses blur in dynamic regions without sacrificing integration time in static areas. Crucially, the Log-Gabor probabilistic events reduce blur RMSE to just 0.72 stops (from the oracle's 1.39) and noise RMSE to 1.50 stops (from 2.24). This yields an overall stability RMSE of 1.28 stops---roughly one photographic stop---compared to the global baseline's 2.0. In physical terms, the 0.72-stop blur limit ensures dynamic regions are overexposed by no more than 1.64x their optimal duration, tightly preserving spatial edges whereas the global shutter suffers errors up to a factor of three.

Probabilistic events recover temporal stability to within roughly one photographic stop, showing that local temporal windows can be estimated reliably from noisy quanta measurements. However, this estimate also depends on the amount of evidence accumulated before inference. We therefore repeat the analysis while varying the number of binary quanta frames summed before inference, $N_{\text{sum}}$, and the incident photon rate. A representative sequence is shown in \cref{fig:analysis_summed_frames_vs_light_level}, with aggregate error distributions in \cref{fig:analysis_summed_frames_vs_light_level_aggregate}. At small $N_{\text{sum}}$, the estimator remains responsive but becomes evidence-limited as light decreases. Increasing $N_{\text{sum}}$ provides stronger local evidence and reduces under-exposure error, but only until the summed window exceeds the local temporal stability of the scene, at which point the error shifts toward motion blur. The aggregate distributions expose this transition: lower light requires more temporal aggregation, whereas excessive aggregation produces the positive-error tail associated with blur. Thus, temporal stability is not only an output of probabilistic events, but also the local limit on useful evidence aggregation.

\subsubsection*{Upper Bound on SNR Advantage}

While the preceding section assessed the practical performance of probabilistic events, we now turn to the intrinsic limits of the per-pixel temporal stability strategy itself. We benchmark our approach against a global shutter baseline---the operating principle of conventional cameras. However, to strictly isolate the benefits of per-pixel adaptation, we simulate this baseline on a quanta image sensor, \ie, using virtual exposures. By computing virtual exposures, we eliminate read-noise penalties and present a favorable performance scenario for conventional global shutters. Such a global strategy is fundamentally constrained by a single exposure time, $\tau_\text{global}$, which must be throttled by the scene's fastest dynamics to preclude motion blur. In contrast, our probabilistic events obviate this compromise by estimating per-pixel temporal stability, which is determined independently by local motion dynamics. Formally, let $v(\vp)$ denote the local velocity magnitude at pixel $\vp$ in pixels per second. Whereas the global exposure is bottlenecked by the maximum velocity in the scene, $\tau_\text{global} \propto (\max_\vp v(\vp))^{-1}$, the probabilistic event camera permits a temporal stability window inversely proportional to the local velocity:
\begin{equation}
    \tau_{\text{ours}}(\vp) \propto v(\vp)^{-1}.
\end{equation}
This pixel-wise adaptation allows static background regions to integrate light for durations significantly exceeding those of dynamic foregrounds, thereby maximizing the Signal-to-Noise Ratio (SNR) without introducing motion blur.

To quantify reconstruction quality, we derive the SNR of the linear intensity estimate. Let $\phi_t(\vp)$ denote the expected photon count incident on pixel $\vp$ during frame $t$. Since our adaptive exposure ensures the scene is effectively static within the integration window, we assume the count is constant, $\phi_t(\vp) \equiv \phi(\vp)$. We apply this same stationarity assumption to the global shutter baseline; however, in \cref{fig:analysis}, we explicitly mark pixels where this assumption fails (i.e., where motion blur occurs) and exclude them from the SNR analysis. Consequently, the sequence of captured binary frames is modeled as a Bernoulli process with parameter $p = 1-e^{-\phi(\vp)}$. For an integration window of $\omega$ binary frames, the Maximum Likelihood Estimator (MLE) of the photon detection probability is $\hat{p} = \frac{1}{\omega} \sum_{i=1}^\omega B_i$. Inverting the Bernoulli response yields the corresponding photon count estimate $\widehat{\phi} = -\ln (1-\hat{p})$. The SNR of the reconstructed linear image is thus defined as:
\begin{equation}
\begin{split}
    \text{SNR}&=20\log_{10}\paren*{\frac{{\widehat{\phi}}}{\sqrt{\text{Var}(\widehat{\phi})}}},\\
    \text{Var}(\widehat{\phi}) &= \text{Var}\paren*{-\ln(1-\hat{p})} = \mathbb{E}[\ln(1-\hat{p})^2] - \mathbb{E}[\ln(1-\hat{p})]^2.
    \label{eq:snr}
\end{split}
\end{equation}

To render the variance term tractable, we approximate the expectations via a Taylor expansion about $a = 0.5$, given by $\mathbb{E}[\ln(X)] \approx \ln(a) + \sum_{n=1}^{\infty}\frac{(-1)^{n-1}}{n a^{n}}\,\mathbb{E}[(X-a)^{n}]$, with an analogous expansion for $\mathbb{E}[\ln(X)^2]$. The central moments $\mathbb{E}[(X-a)^{n}]$ are computed directly from the moment-generating function of the binomial distribution.

This derivation yields a strict theoretical upper bound on SNR for a given photon detection probability $p$ and integration window $\omega$. The validity of this bound rests on two conditions: first, that $p$ remains constant during integration (implying sub-pixel motion), and second, that photon shot noise is the sole noise source. \Cref{fig:analysis}A visualizes the resulting SNR landscape using a twelfth-order Taylor approximation.

We validate these theoretical limits using the VisionSIM dataset~\citemethods{visionsim}, which provides ground truth optical flow and pixel intensities for a variety of realistic indoor scenes. We compare the locally adaptive integration against a conventional global shutter baseline. For the latter, we establish $\tau_\text{global}$ such that the majority of pixels remain blur-free, subject to a hard ceiling of $\nicefrac{1}{1000}$\,s, since most conventional cameras will not realistically attain shorter exposures without additional light. Specifically, we set $\tau_\text{global} = \max(\nicefrac{1}{1000}, (\max_\vp v(\vp))^{-1})$. Conversely, for our method, we optimally set $\tau_{\text{ours}}(\vp)$ to the inverse of the ground truth optical flow $v(\vp)$, ensuring maximal exposure within sub-pixel motion constraints. \Cref{fig:analysis}B illustrates the maximum theoretical SNR gain realized by transitioning from global to pixel-wise optimal exposure. Points falling below the gray line represent pixels that would suffer blur even at the minimum $\tau_\text{global}$ of $\nicefrac{1}{1000}$\,s---a regime where the static assumption of our analysis breaks down. Finally, \cref{fig:analysis}(c) presents the distribution of the relative exposure time increase, $(\tau_{\text{ours}}-\tau_\text{global})/\tau_\text{global}$, across the dataset, while the corresponding improvement in SNR is detailed in \Cref{fig:analysis}D.

\subsection*{Related Work}

\subsubsection*{Quanta Computational Imaging}

Quanta image sensors capture light at the fundamental granularity of individual photon detections, enabling imaging in challenging illumination and motion regimes. These sensors excel simultaneously in low-light \citemethods{ma_2023_wacv,Bruschini2019}, high-dynamic-range \citemethods{inglehighfluxpassive2019,inglepassiveinterphotonimaging2021,liu_2022_wacv}, and high-speed regimes \citemethods{iwabuchi2021,seets_2021_wacv,Wei_2023_ICCV}.
However, raw quanta streams are heavily quantized and dominated by shot noise, making intensity recovery a statistical inference problem rather than a direct readout. Early work developed along two complementary measurement models: SPAD-based quanta image sensors, which exploit temporally precise photon detections and inter-photon arrival times for flux estimation and passive imaging~\citemethods{laurenzis2019single,inglehighfluxpassive2019,inglepassiveinterphotonimaging2021}, and jot-based QIS systems, which record dense binary or multi-bit bit planes that must be aggregated, statistically inverted, or learned from through sensor-aware image formation models~\citemethods{fossum2016quanta,ma2019photon,choi2018image,gnanasambandam2019megapixel,gnanasambandam2020hdr}.

Together, these works represent reconstruction-first quanta computational imaging: photon-level measurements are converted into intensity images by progressively richer models of sensor statistics and image structure, ranging from bit-plane aggregation to color recovery, HDR recovery, and learned reconstruction. Subsequent methods carried this philosophy into dynamic low-light scenes, where recovering images from sparse photon streams requires either learned priors or motion-aware processing \citemethods{chi2020dynamic}. More recent quanta reconstruction methods have further expanded this toolkit, relying on non-causal align-and-merge \citemethods{ma_quanta_2020,ma_2023_wacv}, change-driven spatiotemporal integration~\citemethods{seets_2021_wacv}, self-supervised noise-to-noise restoration~\citemethods{liu2024bit2bit}, diffusion-based generative restoration~\citemethods{garg2026gqir}, and burst-photography-inspired neural networks \citemethods{chennuri2025quanta}.
While highly effective for offline scientific or photographic applications, these reconstruction-first pipelines impose latency costs that fundamentally misalign with real-time robotic perception. Probabilistic events bypass explicit motion compensation, using recursive Bayesian inference to navigate the noise--blur trade-off implicitly and produce actionable representations at kilohertz rates.
Although this work focuses on SPAD-based QIS owing to their high temporal resolution, the formulation is not tied to a specific photon-counting architecture and can be extended to jot-based QIS by replacing the Bernoulli observation model with the truncated-Poisson likelihood.

\subsubsection*{Computer Vision with Quanta Sensors}

Transitioning quanta cameras from offline imagers to real-time perception sensors requires bridging the gap between raw photon arrivals and downstream vision algorithms. Although sketching \citemethods{zhang2024streaming} and selective readout \citemethods{sundar_sodacam_2023,Sundar_2024_CVPR} techniques mitigate the transmission bottlenecks inherent to quanta image sensors, downstream perception remains largely underexplored. In latency-critical applications, perception systems must bypass costly image reconstruction to operate directly on the incoming photon stream.

Prior reconstruction-free methods typically process only limited temporal windows (tens to hundreds of quanta frames) \citemethods{Li_2021_ICCV,gnanasambandam2020image} or assume strictly static scenes \citemethods{Goyal_2021_ICCV}. Eulerian single-photon vision \citemethods{Gupta:2023:Eulerian} forgoes the reconstruction-first approach by passing raw acquisitions through banks of velocity-tuned filters; however, its utility is confined to a restricted set of handcrafted, phase-based algorithms, such as edge detection and motion estimation. More recently, Quanta Neural Networks \citemethods{Sundar_2025_ICCV} offer an efficient computational pathway for processing quanta frames. Yet, despite their compatibility with various image- and video-based networks, they require retraining or fine-tuning of existing architectures---a significant challenge given the scarcity of large-scale quanta datasets.

In contrast, probabilistic events function as lightweight primitives that consume quanta frames at native speeds. This approach yields immediate representations directly compatible with off-the-shelf machine vision models, circumventing both the computational overhead of explicit image reconstruction and the need for domain-specific retraining.

\subsubsection*{Event-Driven Computer Vision}

Event cameras, most commonly the dynamic vision sensor (DVS) \citemethods{lichtsteiner200364x64} and DAVIS \citemethods{davis240}, compress visual information by asynchronously reporting changes in log-intensity. While these sensors offer microsecond latency and high dynamic range, their sparse event streams discard absolute intensity. Consequently, downstream perception requires either bespoke asynchronous algorithms~\citemethods{gallego_2018_cvpr,gehrig2020eklt,benosman2013event,rebecq2016evo,mueggler2017fast} or ill-posed intensity reconstruction to interface with standard computer vision pipelines \citemethods{barua2016direct,zhang2023formulating,rebecq19pami}. Probabilistic events redefine event-based sensing directly on digital photon detections. Rather than relying on deterministic analog thresholds, change detection is formulated as a recursive Bayesian inference problem over photon arrivals. The resulting motion-aware flux dynamically dictates per-pixel exposures to continuously balance the noise--blur trade-off---an approach deeply rooted in the spatially varying exposures of computational photography \citemethods{nayar2000high,nayar2004programmable,zhang2020closed}.

Change-driven processing has also appeared previously in single-photon and quanta imaging, primarily as a mechanism for reconstruction or learned representation formation. Motion-adaptive deblurring with single-photon cameras~\citemethods{seets_2021_wacv} uses changes in photon-count statistics to select local temporal windows, accumulating photons over intervals that avoid crossing motion boundaries before producing a restored intensity image. Generalized Events~\citemethods{Sundar_2024_CVPR} extended the event-camera abstraction to quanta measurements by deriving change events from photon streams, but relied on non-causal backtracking and neural-network decoding, making it unsuitable for ultra-low-latency inference. More recently, the motion-metering layers in Quanta Neural Networks~\citemethods{Sundar_2025_ICCV} introduced adaptive exponential smoothing driven by Bayesian online changepoint detection over Bernoulli photon streams. This purely temporal formulation responds conservatively to abrupt intensity changes and required post-hoc operations, such as minimum filtering of stability times, to suppress excessive blur. While effective as an intermediate layer for neural networks, it does not by itself provide a robust, task-ready representation for low-latency perception.

Probabilistic events take a different view. Rather than converting photon streams into thresholded events, selecting integration windows for reconstruction, or using changepoint estimates only as internal neural-network features, they maintain a causal posterior over temporal stability at each pixel. Apparent scene changes are represented as shifts in this belief state, from which intensity, activity, and uncertainty are jointly derived. The likelihood can also be defined over spatial features rather than isolated pixel measurements, allowing the inference process to exploit local structure and inter-pixel relationships while preserving a causal, low-memory update. This feature-based formulation yields a strictly improved noise--blur trade-off over the purely temporal Bernoulli formulation: it is more responsive to structured scene changes in photon-limited regimes, without requiring explicit motion compensation, offline decoding, or post-hoc blur correction. In addition, moving from Bernoulli photon measurements to a binomial, multi-bit formulation reduces computational overhead while retaining the photon-statistical basis of the model. Together, these spatiotemporal belief updates produce motion-adaptive intensity information while preserving the uncertainty and activity cues that conventional intensity reconstruction and traditional event cameras do not jointly provide.

\subsection*{Additional Experimental Details}

\subsubsection*{Hardware and Sensor Specifications}

We acquire real-world quanta sequences using the SwissSPAD2 sensor~\citemethods{ulku512512spad2019}, operated in half-array mode ($256 \times 512$ pixels), at 96.8~kHz, close to its maximum (binary) frame rate. Results that use this sensor are shown in \cref{fig:method_walkthrough,fig:plug_and_play,fig:qbp_comparison,fig:noise_blur,fig:spatially_varying_denoising}. We also demonstrate scalability to megapixel resolutions in \cref{fig:teaser,fig:color_metering}, which feature quanta sequences captured at 31~kHz using a Bayer color filter array. Note that the conventional capture shown in \cref{fig:teaser} was simulated from quanta sensor data, assuming no readout noise.

\subsubsection*{Baseline Implementations}
To ensure fair comparisons, all learned baselines (QUIVER~\citemethods{chennuri2025quanta}, FastDVDNet~\citemethods{Tassano_2020_CVPR}, NAFNet~\citemethods{chen2022simple}, and Streaming EMA~\citemethods{zhang2024streaming}) were trained on simulated quanta data calibrated to match the SwissSPAD2 sensor's characteristics. We simulated photon arrival events at a temporal resolution equivalent to a 96~kHz frame rate, varying average photon detection probabilities between 0.05 and 0.5 to represent both low-light and ambient illumination conditions.

For the frame-based restoration methods (QUIVER, NAFNet and FastDVDNet), we evaluated test sequences of 4,096 binary frames, aggregating them into non-overlapping 64-frame chunks to construct the effective exposure frames for the networks. For gQIR~\citemethods{garg2026gqir}, we used the publicly available checkpoints released by the authors. For Quanta Burst Photography~\citemethods{ma_quanta_2020}, we used a PyTorch implementation of the original MATLAB code. This implementation additionally skips quanta-frame-level flow interpolation, and instead uses block-based alignment and flow. In our evaluation, this implementation is approximately $50\times$ faster than the original MATLAB code.

For the Streaming EMA baseline, we employed a bank of eight exponential smoothing coefficients, with effective window sizes geometrically spaced from 8 to 1,024 binary frames. To ensure that even the largest window sizes reach steady state, we applied a warmup period of 4,096 binary frames before evaluation; this warmup time was strictly excluded from all throughput calculations.

\subsubsection*{Perception Task Evaluation}

For the perception task evaluation, we used VisionSim rendered videos at 1.6kHz, together with clean intensity, depth, and optical-flow references sampled at 100Hz. Binary quanta measurements were simulated at 51.2kHz, so each 100Hz reference interval corresponded to 512 binary SPAD frames. We varied the photon budget by scaling the rendered intensities to average photon detections per pixel (PPP) of 0.5, 0.05, and 0.025, corresponding to the original light level, $10\times$ fewer photons, and $20\times$ fewer photons, respectively. To test robustness to motion, we evaluated temporal strides of $1\times$, $2\times$, and $5\times$ relative to the original VisionSim sequence.

For each reference time, we constructed the compared photon-stream representations from the simulated 51.2kHz binary measurements. The virtual-exposure baseline summed 32 binary frames, corresponding to a 0.625ms exposure. QBP operated on 32-frame count images and was run on a full 8,192-binary-frame horizon, corresponding to 160ms of quanta data; internally, dense outputs were produced using centered five-count-frame alignment and merge windows. QUIVER operated on 64-frame count images and used five past and five future count frames in addition to the target frame, yielding an 11-frame input window spanning 704 binary frames, or 13.75ms. Our probabilistic event representation also used 32-frame count images, corresponding to 0.625~ms samples.

Image fidelity was measured against the clean rendered frame using peak signal-to-noise ratio (PSNR) and structural similarity index measure (SSIM). The reference intensity was assigned infinite PSNR and SSIM of 1.0 by convention.

For monocular depth, DepthAnything-v2 predicts relative inverse depth. Let $r_\vp$ denote the raw predicted inverse-depth value and let $d_\vp$ denote the metric ground-truth depth at pixel $\vp \in \Omega$, where $\Omega$ is the set of evaluated pixels. Before computing depth metrics, we affine-align the prediction to ground-truth inverse depth and convert it to metric depth:
\begin{equation}
\begin{split}
    a^\star,b^\star
    &= \arg\min_{a,b}\sum_{\vp\in\Omega}
    \left(a r_\vp + b - \frac{1}{d_\vp}\right)^2, \\
    \hat d_\vp
    &= \mathrm{clip}\left(
    \frac{1}{\max(a^\star r_\vp+b^\star,\epsilon)},
    d_{\min}, d_{\max}
    \right).
\end{split}
\end{equation}
where $\epsilon$ is a small positive constant for numerical stability, and $d_{\min}$ and $d_{\max}$ are the nearest and farthest valid ground-truth depths in the evaluated frame, respectively. The clipping step is used only to keep affine-aligned relative-depth predictions within the valid scene depth range. We then report relative depth error,
\begin{equation}
    \mathrm{RelErr} =
    \frac{1}{|\Omega|}\sum_{\vp\in\Omega}
    \frac{|\hat d_\vp-d_\vp|}{d_\vp},
\end{equation}
and threshold accuracy,
\begin{equation}
    \mathrm{ThrAcc} =
    \frac{1}{|\Omega|}\sum_{\vp\in\Omega}
    \mathbf{1}\left[
    \max\left(\frac{\hat d_\vp}{d_\vp},
    \frac{d_\vp}{\hat d_\vp}\right)\leq 1.25
    \right].
\end{equation}

For optical flow, let $\hat{\mathbf{u}}\vp$ and $\mathbf{u}\vp$ denote the predicted and ground-truth two-dimensional flow vectors at pixel $\vp$, respectively. We report endpoint error,
\begin{equation}
    \mathrm{EndErr} =
    \frac{1}{|\Omega|}\sum_{\vp\in\Omega}
    |\hat{\mathbf{u}}\vp-\mathbf{u}\vp|2,
\end{equation}
and bad-pixel rate,
\begin{equation}
    \mathrm{BadPix} =
    \frac{1}{|\Omega|}\sum{\vp\in\Omega}
    \mathbf{1}\left[
    |\hat{\mathbf{u}}\vp-\mathbf{u}\vp|_2 > 3
    \right].
\end{equation}

\subsection*{Extended Discussion}

\subsubsection*{Quanta Sensor Non-Idealities}

The observation models developed in this work assume idealized photon-counting measurements. Practical SPAD-based quanta image sensors deviate from this assumption through detector dead time, dark counts, afterpulsing, optical crosstalk, saturation, and pile-up effects. These effects perturb the statistics of the measured photon stream, particularly near the operating limits of the sensor. However, all real-data experiments presented in this work are performed using physical quanta image sensors and therefore inherently include these non-idealities. The stable operation of probabilistic events across a wide range of illumination levels and motion regimes, using data acquired from two distinct quanta sensors suggests that the framework is not unduly sensitive to moderate departures from the assumed Bernoulli model. Moreover, these non-idealities have become increasingly controlled in modern imaging SPAD arrays: SwissSPAD2 reports minimum dark-count-rate densities of approximately $0.18$ cps/$\mu$m$^2$~\citepmethods{ulku512512spad2019}, while recent charge-focusing SPAD architectures report DCR densities near $0.015$ cps/$\mu$m$^2$ at room temperature~\citepmethods{morimoto2020charge}. Prior characterization studies further show that practical photon-counting response curves, including deviations induced by dead time and recharge behavior, can be measured and modeled with reasonable accuracy~\citepmethods{antolovic:18}.

Broadly, the Bernoulli likelihood should be viewed as a tractable observation model rather than a complete physical description of detector behavior. Sensor non-idealities may alter the effective photon statistics or reduce posterior confidence near saturation or in extremely dark conditions, but they do not change the recursive structure of the Bayesian update. In the photon-limited operating regimes considered here, photon shot noise and scene motion remain the dominant factors governing perception quality. If a calibrated detector model is available, the likelihood can be replaced with a richer observation model incorporating dark counts, dead time, saturation, or afterpulsing without changing the overall probabilistic-event framework. Finally, if such effects become limiting, another possible approach would be to incorporate more robust Bayesian change-detection variants based on heavy-tailed likelihoods or generalized Bayesian updates, further reducing sensitivity to isolated detector artifacts and outlier photon counts~\citepmethods{knoblauch2018doubly,altamirano2023robust,fearnhead2019changepoint}.

%% file: figures/reconstruction_results.tex
\begin{figure*}[t]
    \centering
    \includegraphics[width=\textwidth]{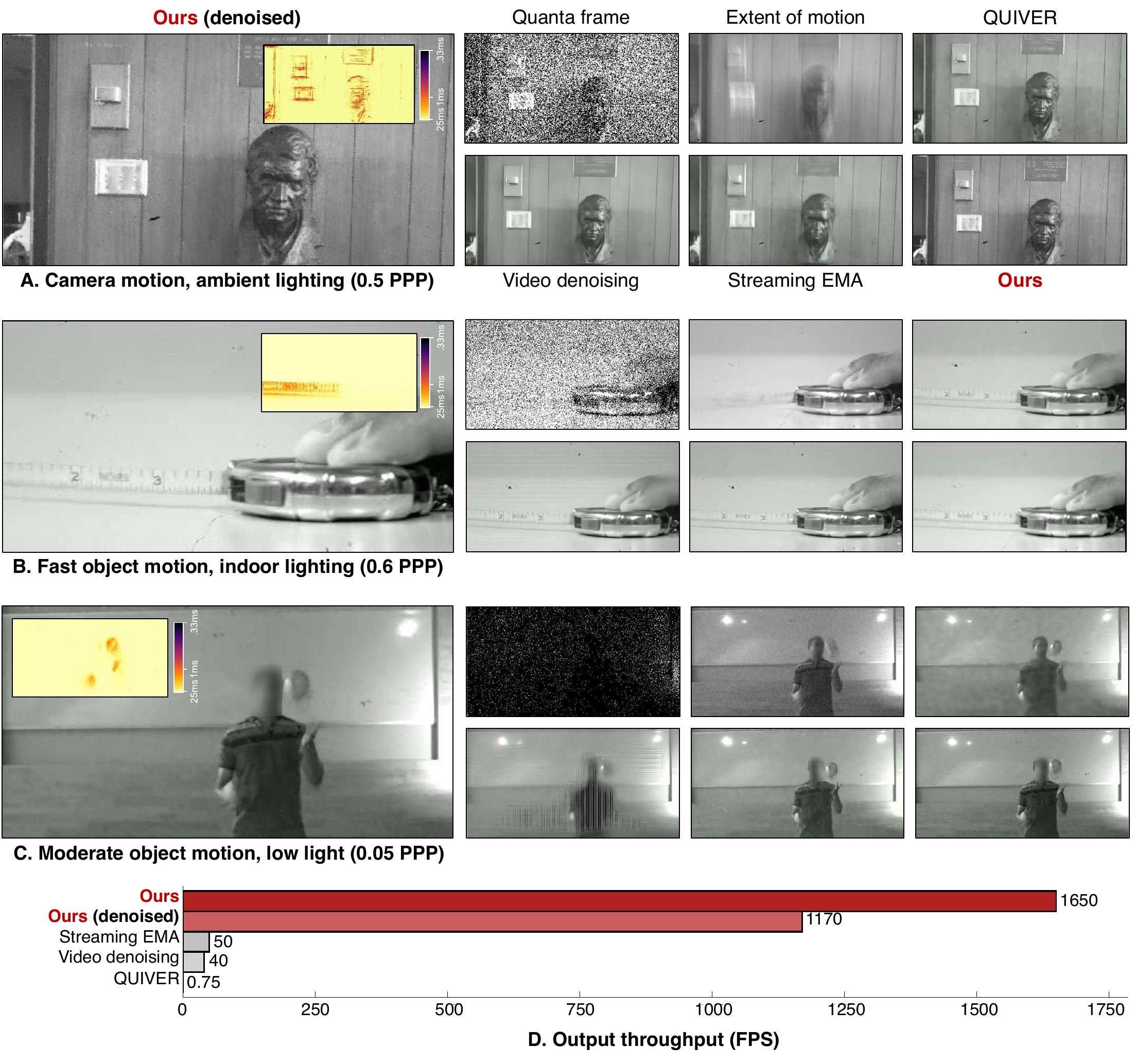}
    \vspace{-0.2in}
    \caption{\textbf{Comparing probabilistic events' motion-aware aggregation to quanta-image restoration.}
    We compare against burst restoration using QUIVER~\protect\citepmethods{chennuri2025quanta}, video denoising using FastDVDnet~\protect\citepmethods{Tassano_2020_CVPR}, and online synthesis using Streaming EMA~\protect\citepmethods{zhang2024streaming}.
    We show three imaging scenarios: \textbf{(A)} camera motion in ambient lighting, \textbf{(B)} fast object motion in indoor lighting, and \textbf{(C)} moderate object motion in low light. \textit{Faces have been blurred to protect privacy.}
    We also show a single binary frame of the quanta sensor acquisition and the extent of motion in 30~ms.
    Our approach, despite being computationally much lighter, provides output fidelity comparable to, and in certain cases, better than, the considered baselines.
    Insets show the temporal stability maps computed by probabilistic events: dynamic regions are assigned shorter windows to minimize motion blur.
    In challenging conditions involving fast motion or low light, probabilistic events better preserve high-frequency details (e.g., numbers and markings on the measure tape) that are oversmoothed by learned baselines.
    \textbf{(D)} Our approach (red bars) is the only method capable of exceeding 1,000~FPS, operating $33$--$2200\times$ faster than the considered baselines; we illustrate this using a stacked bar chart. All runtime measurements assume desktop-grade computing resources: an Intel Xeon Silver CPU fitted with an NVIDIA 4090 GPU accelerator.}
    \label{fig:reconstruction_results}
\end{figure*}

%% file: tables/i2k_metrics.tex
\begin{table}[h]
    \centering
    \caption{\textbf{Image quality comparisons on the i2k high-speed dataset~\protect\citepmethods{chennuri2025quanta}.}}
    \begin{tabular}{@{} l r rr rr @{}}
        \toprule
        \multirowcell{2}[-1pt]{Method}  & \multirowcell{2}[-1pt]{Output\\ FPS} & 
        \multicolumn{2}{c}{PPP=$0.05$} &
        \multicolumn{2}{c}{PPP=$0.5$} \\
        \cmidrule(lr){3-4} \cmidrule(lr){5-6}
        & & %
        PSNR$\uparrow$ (dB) & SSIM$\uparrow$ & 
        PSNR$\uparrow$ (dB) & SSIM$\uparrow$  
        \\
        \midrule
        \multicolumn{6}{c}{\textit{Baselines}} \\ 
        \midrule
        Virtual exposure & 10000 &
        11.0 & 0.068 &
        19.3 & 0.212  \\

        Exposure stack~\citepmethods{zhang2024streaming} & \textbf{80} &
        28.8 & 0.809 &
        30.6 & 0.869  \\

        FastDVDNet~\citepmethods{Tassano_2020_CVPR} & 50 &
        24.1 & 0.528 &
        {33.5} & {0.896} \\

        NAFNet~\citepmethods{chen2022simple} & 20 &
        22.2 & 0.448 &
        27.0 & 0.774 \\

        QUIVER~\citepmethods{chennuri2025quanta} & 0.75 &
        \textbf{29.7} & \textbf{0.804} &
        \textbf{33.5} & 0.868  \\

        QBP~\citepmethods{ma_quanta_2020} & 0.5 &
        29.4 & 0.649 &
        31.2 & \textbf{0.900}  \\

        gQIR~\citepmethods{garg2026gqir} & 0.3 &
        27.4 & 0.780 &
        27.5 & 0.804  \\

        \midrule
        \multicolumn{6}{c}{\textit{Probabilistic event camera (ours)}} \\ 
        \midrule
        Gradient & 4900 &
        26.1 & 0.606 &
        30.5 & 0.835  \\

        \textit{with denoising} & 2200 &
        27.1 & 0.725 &
        30.86 & 0.869  \\

        Log-Gabor & 1650 &
        27.6 & 0.676 &
        32.3 & 0.866  \\

        \textit{with denoising} & \textbf{1170} &
        \textbf{29.0} & \textbf{0.820} &
        \textbf{32.9} & \textbf{0.884}  \\       
        \bottomrule
    \end{tabular}
    \vspace{0.1in}
    \footnotesize
    \textbf{Table legend:} We benchmark image-quality metrics (PSNR and SSIM) and output throughput (FPS) at low ($0.05$ PPP) and moderate ($0.5$ PPP) light levels. While virtual exposures are fast ($10,000$ FPS), they lack denoising capabilities. Conversely, neural methods like QUIVER achieve high quality but are too slow for real-time applications ($<1$ FPS). The probabilistic event camera strikes an optimal balance: our Log-Gabor instantiation with denoising approaches state-of-the-art quality (within $0.7$ dB of QUIVER) while maintaining kilohertz-rate throughput ($>1,000$ FPS), making it a high-fidelity solution suitable for low-latency vision. \textbf{Bold} entries denote the top-performing method within the baseline and proposed categories, respectively.
    \label{tab:i2k_metrics}
\end{table}

%% file: figures/reconstruction_results_qualitative.tex
\begin{figure*}[t]
    \centering
    \includegraphics[width=\textwidth]{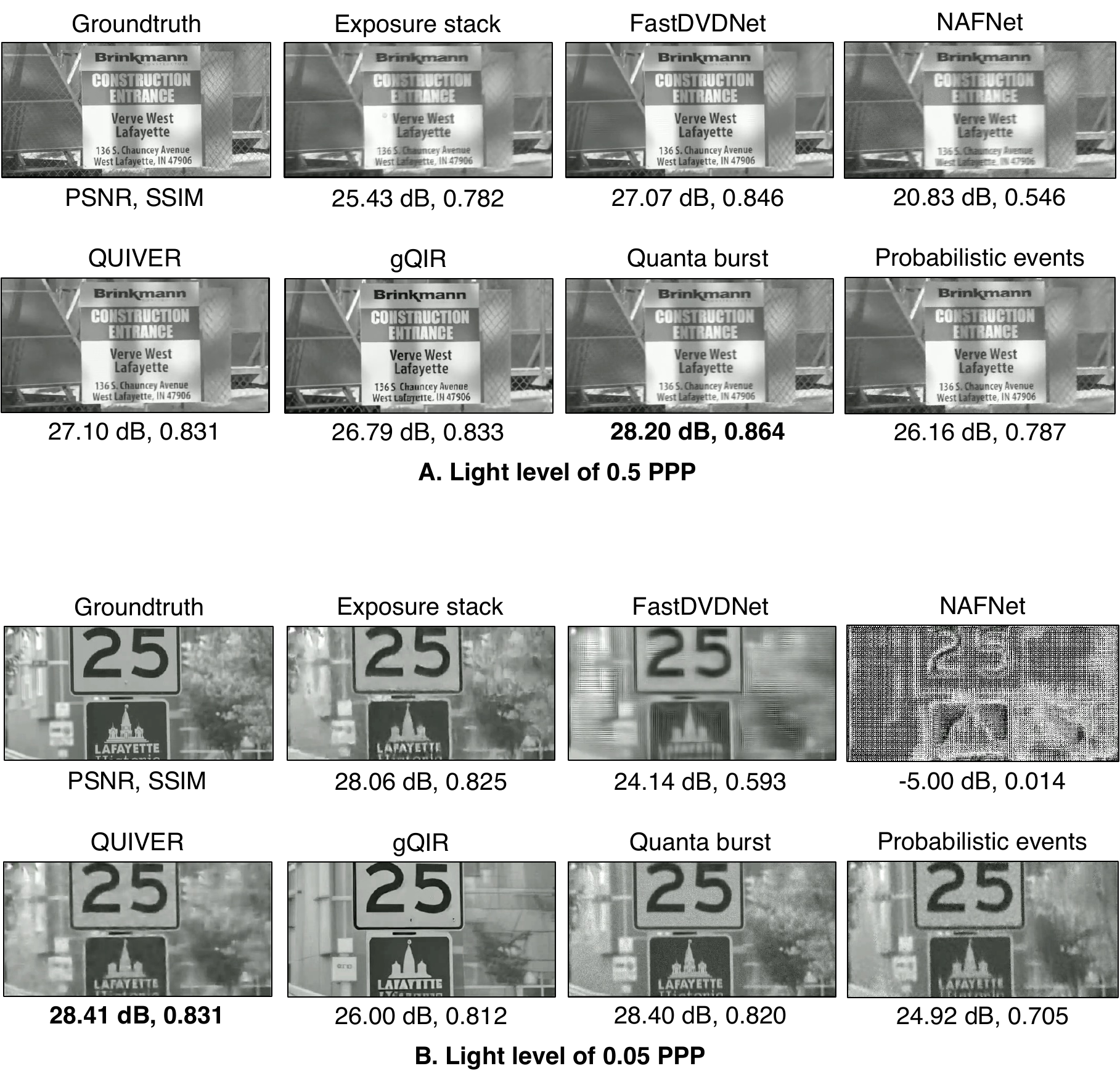}
    \vspace{-0.2in}
    \caption{\textbf{Qualitative image-quality comparisons on the i2k high-speed dataset.}
    We show two representative examples from the i2k benchmark~\protect\citepmethods{chennuri2025quanta}, at moderate light ($0.5$ PPP; A) and low light ($0.05$ PPP; B), complementing the aggregate PSNR, SSIM, and throughput results in \cref{tab:i2k_metrics}.
    Each example compares the clean ground truth with exposure stacks~\protect\citepmethods{zhang2024streaming}, FastDVDNet~\protect\citepmethods{Tassano_2020_CVPR}, NAFNet~\protect\citepmethods{chen2022simple}, QUIVER~\protect\citepmethods{chennuri2025quanta}, gQIR~\protect\citepmethods{garg2026gqir}, Quanta Burst Photography~\protect\citepmethods{ma_quanta_2020}, and the proposed probabilistic-event representation.
    The image- and video-denoising baselines, NAFNet and FastDVDNet, rely on limited temporal context and degrade markedly at very low photon counts, whereas quanta-specific restoration methods exploit the binary photon stream more effectively.
    Among these methods, QBP produces visually strong reconstructions, while QUIVER benefits from end-to-end training and achieves competitive image-quality metrics.
    Probabilistic events provide comparable PSNR and SSIM to state-of-the-art quanta restoration methods at moderate light while operating up to four orders of magnitude faster in this benchmark.
    At low light, reconstruction-first methods recover higher-SNR images, but probabilistic events preserve sufficient scene structure for downstream perceptual inference at kilohertz output rates, as illustrated in \cref{fig:plug_and_play} and quantified for depth estimation and optical flow in \cref{tab:visionsim_depth_flow_metrics}.
    Values below each image report PSNR and SSIM relative to the ground truth.}    \label{fig:reconstruction_results_qualitative}
\end{figure*}

%% file: tables/visionsim_depth_flow_metrics.tex
\begin{table*}[t]
    \centering
    \caption{\textbf{Evaluating task performance on the VisionSim dataset.}
    }
    \resizebox{\textwidth}{!}{%
    \begin{tabular}{@{} cc l rr rr rr @{}}
        \toprule
        \multirowcell{2}{Avg. PPP} & \multirowcell{2}{Motion} & \multirowcell{2}{Method} &
        \multicolumn{2}{c}{Image quality} &
        \multicolumn{2}{c}{Monocular depth (DepthAnything-v2)} &
        \multicolumn{2}{c}{Optical flow (RAFT)} \\
        \cmidrule(lr){4-5} \cmidrule(lr){6-7} \cmidrule(lr){8-9}
        & & & PSNR$\uparrow$ & SSIM$\uparrow$ & Relative error$\downarrow$ & Threshold accuracy$\uparrow$ & Endpoint error$\downarrow$ & Bad-pixel rate$\downarrow$ \\
        \midrule
        \multirow{15}{*}{\makecell{0.025\\20$\times$ fewer\\photons}} & \multirow{5}{*}{1$\times$} & Reference intensity & $\infty$ & 1.000 & 0.067 & 0.940 & 1.08 & 0.053 \\
         &  & Virtual exposure & 9.01 & 0.273 & 0.184 & 0.704 & 21.15 & 0.516 \\
         &  & QBP & 25.93 & 0.945 & 0.086 & 0.916 & \textbf{0.97} & \textbf{0.069} \\
         &  & QUIVER & 25.38 & 0.937 & 0.159 & 0.763 & 5.50 & 0.202 \\
         &  & Probabilistic events & \textbf{29.04} & \textbf{0.970} & \textbf{0.081} & \textbf{0.917} & 1.37 & 0.122 \\
        \cmidrule(lr){2-9}
         & \multirow{5}{*}{2$\times$} & Reference intensity & $\infty$ & 1.000 & 0.060 & 0.952 & 1.12 & 0.084 \\
         &  & Virtual exposure & 8.99 & 0.272 & 0.177 & 0.720 & 23.83 & 0.566 \\
         &  & QBP & 25.44 & 0.933 & 7.541 & \textbf{0.918} & \textbf{1.67} & \textbf{0.130} \\
         &  & QUIVER & 25.41 & 0.935 & 0.151 & 0.781 & 5.88 & 0.262 \\
         &  & Probabilistic events & \textbf{27.53} & \textbf{0.956} & \textbf{0.095} & 0.911 & 3.47 & 0.303 \\
        \cmidrule(lr){2-9}
         & \multirow{5}{*}{5$\times$} & Reference intensity & $\infty$ & 1.000 & 0.054 & 0.960 & 5.14 & 0.262 \\
         &  & Virtual exposure & 9.19 & 0.276 & 0.187 & 0.684 & 31.78 & 0.780 \\
         &  & QBP & 24.65 & 0.924 & \textbf{0.099} & \textbf{0.887} & \textbf{7.07} & 0.475 \\
         &  & QUIVER & 25.64 & \textbf{0.939} & 0.153 & 0.767 & 8.38 & \textbf{0.446} \\
         &  & Probabilistic events & \textbf{25.85} & 0.938 & 0.113 & 0.870 & 14.38 & 0.650 \\
        \midrule
        \multirow{15}{*}{\makecell{0.05\\10$\times$ fewer\\photons}} & \multirow{5}{*}{1$\times$} & Reference intensity & $\infty$ & 1.000 & 0.067 & 0.940 & 1.08 & 0.053 \\
         &  & Virtual exposure & 11.07 & 0.402 & 0.175 & 0.711 & 6.38 & 0.213 \\
         &  & QBP & 28.95 & 0.971 & \textbf{0.076} & \textbf{0.933} & \textbf{1.15} & \textbf{0.062} \\
         &  & QUIVER & 28.07 & 0.964 & 0.118 & 0.853 & 4.47 & 0.159 \\
         &  & Probabilistic events & \textbf{30.74} & \textbf{0.980} & 0.077 & 0.925 & 1.19 & 0.101 \\
        \cmidrule(lr){2-9}
         & \multirow{5}{*}{2$\times$} & Reference intensity & $\infty$ & 1.000 & 0.060 & 0.952 & 1.12 & 0.084 \\
         &  & Virtual exposure & 11.06 & 0.400 & 0.187 & 0.723 & 8.06 & 0.268 \\
         &  & QBP & 28.18 & 0.961 & 0.077 & \textbf{0.942} & \textbf{1.66} & \textbf{0.132} \\
         &  & QUIVER & 28.12 & 0.963 & 0.108 & 0.873 & 4.51 & 0.200 \\
         &  & Probabilistic events & \textbf{29.47} & \textbf{0.972} & \textbf{0.076} & 0.940 & 2.82 & 0.236 \\
        \cmidrule(lr){2-9}
         & \multirow{5}{*}{5$\times$} & Reference intensity & $\infty$ & 1.000 & 0.054 & 0.960 & 5.14 & 0.262 \\
         &  & Virtual exposure & 11.25 & 0.404 & 0.172 & 0.709 & 10.44 & 0.614 \\
         &  & QBP & 26.80 & 0.950 & \textbf{0.081} & \textbf{0.919} & \textbf{6.16} & 0.406 \\
         &  & QUIVER & \textbf{28.38} & \textbf{0.966} & 0.107 & 0.869 & 6.83 & \textbf{0.404} \\
         &  & Probabilistic events & 28.20 & 0.965 & 0.087 & 0.914 & 10.38 & 0.519 \\
        \midrule
        \multirow{15}{*}{\makecell{0.5\\Original\\light level}} & \multirow{5}{*}{1$\times$} & Reference intensity & $\infty$ & 1.000 & 0.067 & 0.940 & 1.08 & 0.053 \\
         &  & Virtual exposure & 22.64 & 0.898 & 0.078 & 0.933 & 0.88 & 0.056 \\
         &  & QBP & \textbf{37.85} & \textbf{0.996} & \textbf{0.066} & \textbf{0.942} & \textbf{0.65} & \textbf{0.046} \\
         &  & QUIVER & 35.54 & 0.994 & 0.071 & 0.935 & 4.99 & 0.169 \\
         &  & Probabilistic events & 37.15 & 0.995 & 0.069 & 0.939 & 0.80 & 0.067 \\
        \cmidrule(lr){2-9}
         & \multirow{5}{*}{2$\times$} & Reference intensity & $\infty$ & 1.000 & 0.060 & 0.952 & 1.12 & 0.084 \\
         &  & Virtual exposure & 22.64 & 0.893 & 0.073 & 0.941 & \textbf{1.31} & \textbf{0.073} \\
         &  & QBP & 35.43 & 0.990 & \textbf{0.061} & 0.952 & 1.48 & 0.119 \\
         &  & QUIVER & 35.62 & \textbf{0.993} & 0.066 & 0.948 & 5.30 & 0.193 \\
         &  & Probabilistic events & \textbf{36.32} & 0.993 & 0.062 & \textbf{0.953} & 1.48 & 0.105 \\
        \cmidrule(lr){2-9}
         & \multirow{5}{*}{5$\times$} & Reference intensity & $\infty$ & 1.000 & 0.054 & 0.960 & 5.14 & 0.262 \\
         &  & Virtual exposure & 22.73 & 0.895 & 0.068 & 0.937 & 4.22 & 0.295 \\
         &  & QBP & 31.31 & 0.976 & 0.062 & 0.948 & 5.90 & 0.346 \\
         &  & QUIVER & \textbf{35.60} & \textbf{0.993} & 0.063 & 0.946 & 7.53 & 0.368 \\
         &  & Probabilistic events & 35.36 & 0.992 & \textbf{0.057} & \textbf{0.956} & \textbf{4.17} & \textbf{0.276} \\
        \bottomrule
    \end{tabular}
    }\\
    \vspace{0.1in}
    \begin{minipage}{\textwidth}
    \footnotesize
    \raggedright
    \textbf{Table legend:} We evaluate photon-stream representations on VisionSim, where clean images, depth, and optical flow are available as references.
    We compare reference intensity frames, fixed 0.625~ms virtual exposures, QBP and QUIVER reconstructions, and probabilistic events using the Log-Gabor variant.
    The 0.625~ms exposure was chosen to prefer noise over blur, since pretrained models such as DepthAnything-v2 are often more robust to noise than motion blur.
    Each representation is evaluated for image fidelity using PSNR and SSIM, monocular depth using DepthAnything-v2, and optical flow using RAFT.
    Probabilistic events provide inference quality comparable to QBP and QUIVER in the original and $10\times$ fewer-photon settings across the tested motion extents, while avoiding full image restoration.
    At $20\times$ fewer photons, performance degrades gracefully: probabilistic events do not uniformly match reconstruction-first methods, especially for large-motion optical flow, but remain substantially better than fixed-window virtual exposures.
    PPP denotes average photon detections per pixel; motion is controlled by temporal stride.
    For depth, relative error is the average relative depth error, and threshold accuracy is the fraction of pixels within 25\% of the ground truth.
    For optical flow, endpoint error is the average Euclidean error between predicted and ground-truth flow vectors, and bad-pixel rate is the fraction of pixels whose endpoint error exceeds three pixels.
    Bold entries denote the best non-reference method within each condition.
    Experimental details are provided in the Methods. See \cref{fig:visionsim_depth_flow} for qualitative examples.
    \end{minipage}
    \label{tab:visionsim_depth_flow_metrics}
\end{table*}

%% file: figures/visionsim_depth_flow.tex
\begin{figure*}[t]
    \centering
    \includegraphics[width=\textwidth]{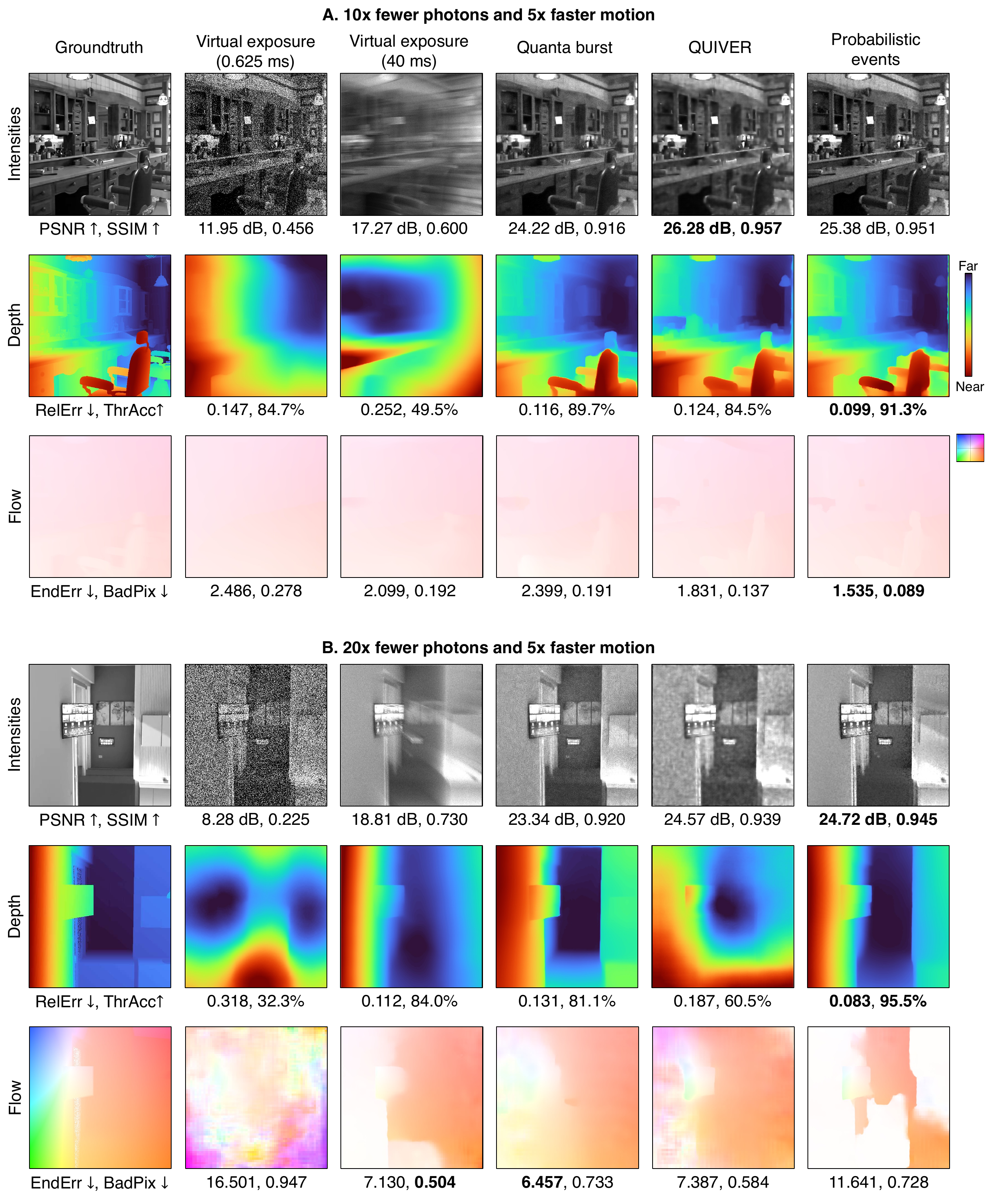}
    \vspace{-0.2in}
\caption{\textbf{Controlled depth and flow evaluation on the VisionSim dataset.}
Representative examples from a controlled task-performance study at $10\times$--$20\times$ fewer photons and $5\times$ faster motion. Columns compare fixed-window virtual exposures, reconstruction-first perception using quanta-burst and QUIVER reconstructions, and probabilistic events. Rows in panels \textbf{A} and \textbf{B} show the input intensity representation, predicted monocular depth, and predicted optical flow. Short virtual exposures preserve motion but are photon-starved; long exposures reduce noise but introduce motion blur. Reconstruction-first methods are strong align-and-merge baselines, although residual artifacts can propagate to downstream estimates. Probabilistic events use causal, motion-adaptive aggregation rather than full reconstruction, yet match reconstruction-first depth and flow at $10\times$ fewer photons. At $20\times$ fewer photons, depth remains robust, while optical flow is more sensitive to residual noise and extreme motion.}
    \label{fig:visionsim_depth_flow}
\end{figure*}

%% file: figures/noise_blur.tex
\begin{figure}[t]
    \centering
    \includegraphics[width=\linewidth]{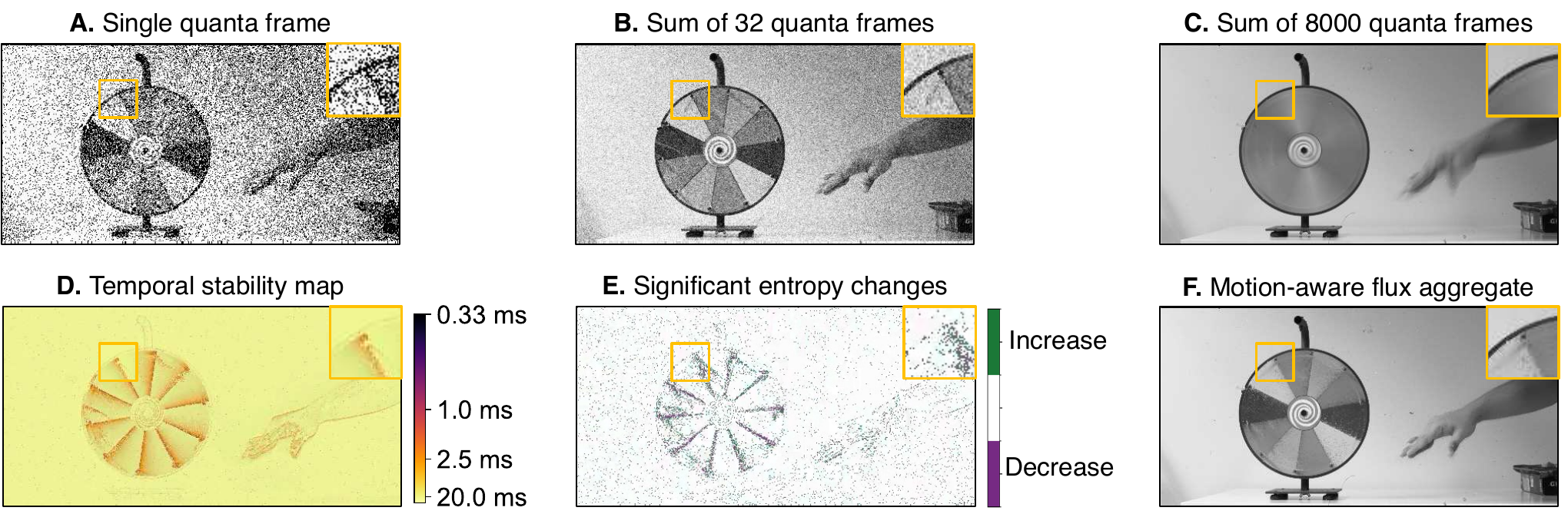}
    \vspace{-0.2in}
    \caption{\textbf{Balancing noise and blur with probabilistic events.}
    Standard fixed integration windows (\cref{eq:sum_image}) impose a rigid global trade-off: a single raw quanta frame (A) or short integration (B, 32 frames) results in significant noise, whereas extended integration (C, 8000 frames) induces motion blur.
    In contrast, probabilistic events utilize adaptive exponential smoothing driven by estimated temporal stability. Depicted are the temporal stability maps (D), shifts in distribution entropy (E, green and purple indicating increase and decrease, respectively), and the motion-aware flux aggregate (F). While this Bernoulli formulation mediates the noise--blur trade-off more effectively than fixed windows, it neglects spatial information; consequently, it remains less responsive to abrupt changes, as evidenced by the blur in the insets.}
    \label{fig:noise_blur}
\end{figure}

%% file: figures/using_spatial_info.tex
\begin{figure*}[t]
    \centering
    \includegraphics[width=\textwidth]{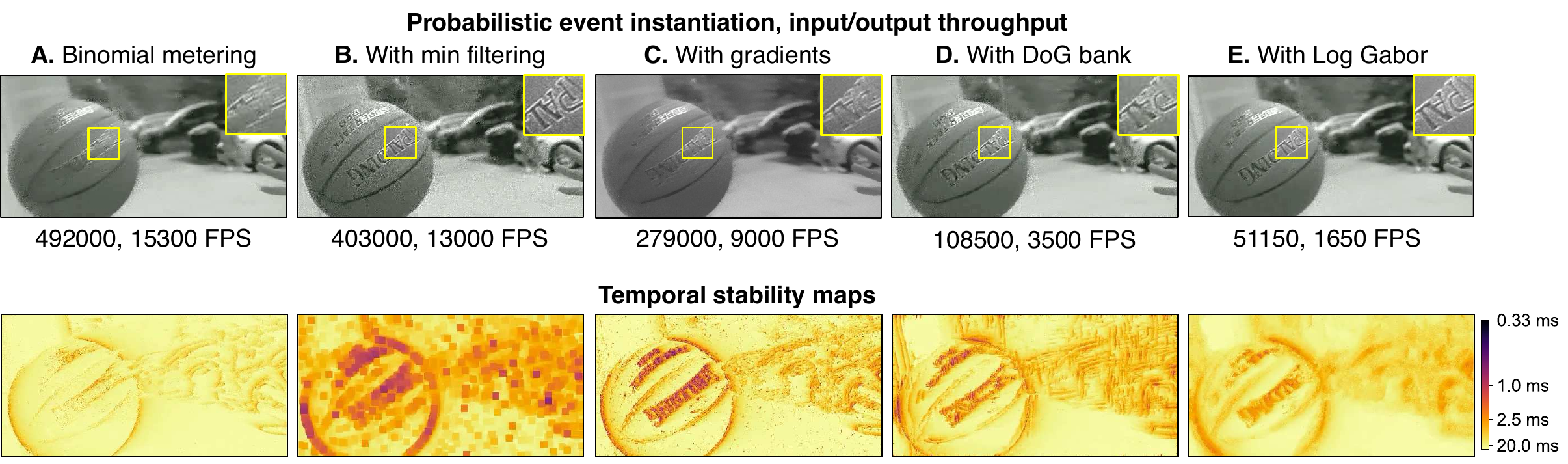}
    \vspace{-0.2in}
    \caption{\textbf{Incorporating spatial information in estimating temporal stability maps.} \textbf{(A)} On a sequence simulated from the i2k high-speed video dataset, the purely pixel-wise binomial instantiation treats locations independently, resulting in laggard responses (increased motion blur). \textbf{(B)} Minimum filtering reduces blur by enforcing smaller temporal stability values if any pixel in a local patch detects change, but does so at the cost of excess noise, as it remains agnostic to scene content. \textbf{(C--E)} In contrast, by explicitly modeling linear spatial features alongside binomial counts, we selectively increase pixel responsivity in textured regions while maintaining robust temporal denoising elsewhere. Features attuned to scale and orientation (e.g., Log-Gabor, E) offer superior selectivity, while simpler features (e.g., gradients, C) provide a computationally lighter alternative. For each instantiation, we show the motion-aware flux aggregate and the corresponding temporal stability map.}
    \label{fig:using_spatial_info}
\end{figure*}

%% file: figures/metering_runtime.tex
\begin{figure}[t]
    \centering
    \includegraphics[width=\linewidth]{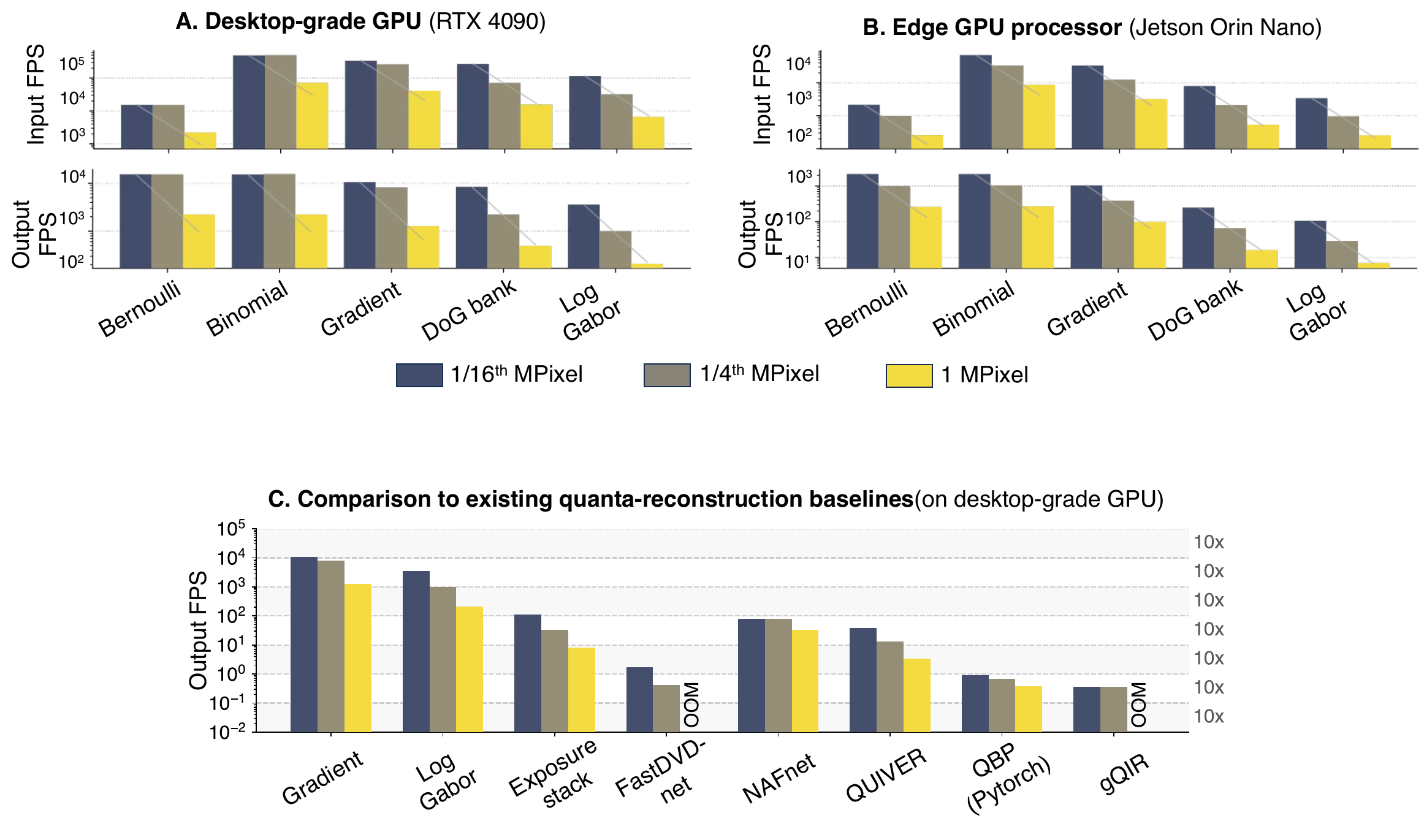}
    \vspace{-0.2in}
    \caption{\textbf{Benchmarking input and output throughput on a desktop-grade GPU (RTX 4090) and an embedded edge platform (Jetson Orin Nano).} Input throughput denotes the binary quanta-frame rate processed by the method, while output throughput denotes the rate at which probabilistic-event representations or reconstructed frames are emitted. \textbf{(A)} On the RTX 4090, the Gradient (\textit{fast} variant) and Log-Gabor (\textit{robust} variant) configurations support a megapixel quanta camera at 41{,}000 and 6{,}700 qFPS, respectively, corresponding to 1{,}286 and 209 FPS output throughput. \textbf{(B)} On the Jetson Orin Nano, the Binomial and Gradient variants sustain $\approx 8{,}700$ and $\approx 3{,}200$ qFPS at 1~MPixel ($\approx 270$ and $\approx 100$ FPS output throughput), confirming viability on power-constrained edge platforms. \textbf{(C)} On the RTX 4090, reconstruction-first quanta baselines such as QBP~\protect\citepmethods{ma_quanta_2020}, QUIVER~\protect\citepmethods{chennuri2025quanta}, and gQIR~\protect\citepmethods{garg2026gqir} operate at sub-Hz to low-Hz output rates. OOM denotes out of (GPU) memory here. Probabilistic events run up to four orders of magnitude faster across the tested resolutions and variants.}
\label{fig:metering_runtime}
\end{figure}

%% file: figures/spatially_varying_denoising.tex
\begin{figure}[t]
    \centering
    \includegraphics[width=\linewidth]{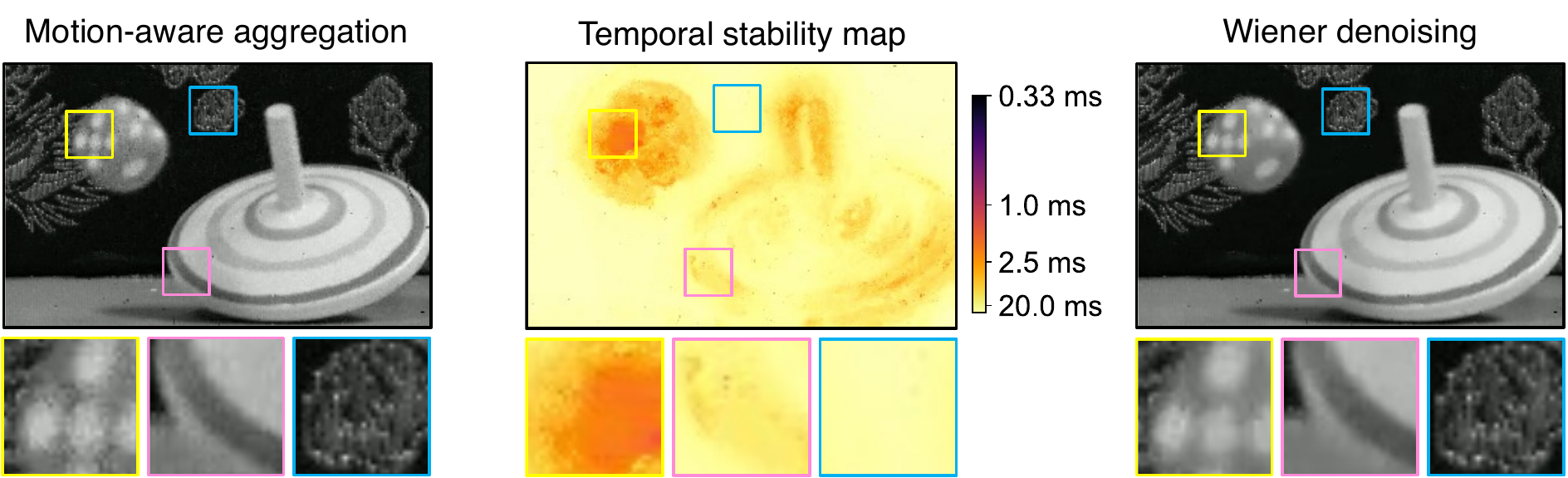}
    \vspace{-0.1in}
    \caption{\textbf{Denoising using spatially-varying Wiener filtering.} We use the temporal stability maps to denoise based on the amount of integration: lesser spatial denoising in regions with strong temporal integration. Going from insets on the left to the right, the amount of motion decreases, the exposure times are larger, and we perform lesser spatial denoising.
    }
    \label{fig:spatially_varying_denoising}
\end{figure}

%% file: figures/rate_distortion_events_vs_video_codec.tex
\begin{figure*}[t]
    \centering
    \includegraphics[width=\textwidth]{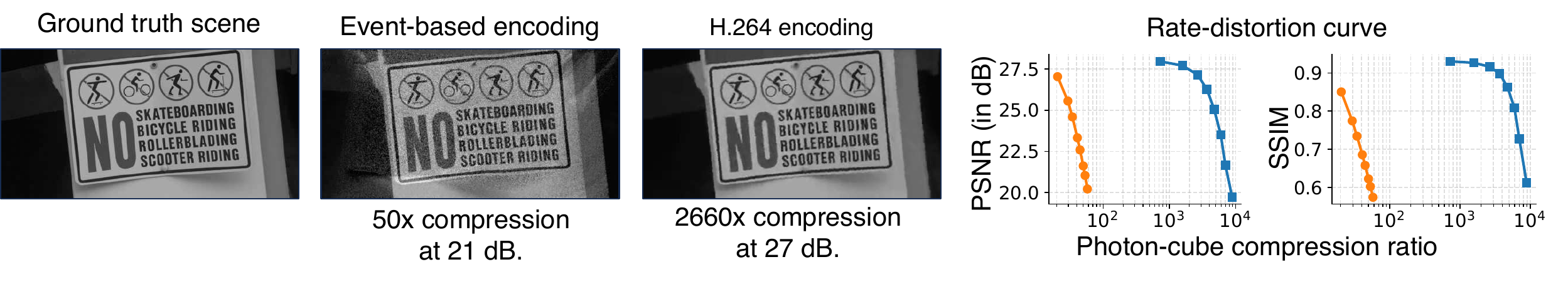}
    \vspace{-0.2in}
    \caption{\textbf{Rate-distortion comparison between event-based and video codec compression.} We compare the compression efficiency of transmitting the motion-aware flux aggregate (after the proposed spatial denoising), using a sparse event-based scheme versus a GPU-accelerated H.264 video codec. The ground truth scene is shown alongside examples of both encoding methods at high compression ratios. The event-based encoding (threshold 0.06) achieves a 50× compression ratio at 21 dB PSNR but suffers from distinct laggard artifacts, visible as ghosting on the signboard during camera panning. In contrast, the H.264 encoding (constant rate factor, CRF, 32) achieves a significantly higher 2660$\times$ compression ratio at 27 dB PSNR. While the video codec exhibits compression artifacts at this extreme CRF setting, such as the loss of high-frequency details leading to a slightly washed-out appearance, it preserves structural integrity far better than the event-based encoding.}    \label{fig:rate_distortion_events_vs_video_codec}
\end{figure*}

%% file: figures/color_metering.tex
\begin{figure}[t]
    \centering
    \includegraphics[width=\columnwidth]{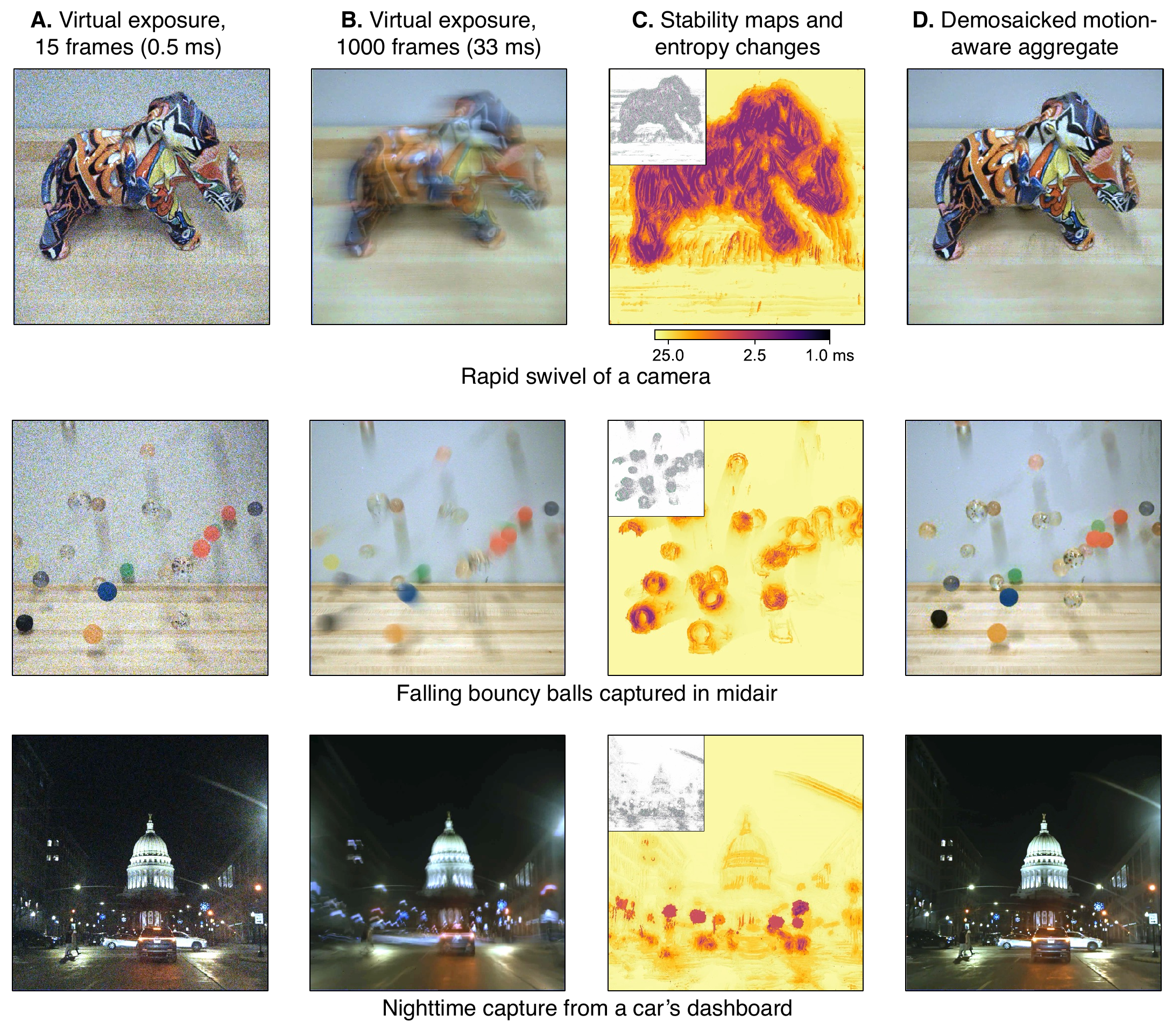}
    \vspace{-0.2in}
    \caption{\textbf{Probabilistic event representations in color.} 
    Color imaging with quanta sensors requires addressing the coupled challenges of demosaicking and denoising. Specifically, we must aggregate sufficient photons to facilitate reliable debayering without incurring motion blur.
    We demonstrate this noise--blur trade-off on \textbf{1-megapixel color quanta acquisition} across three scenes: rapid camera swiveling, high-speed falling objects, and nighttime driving.
    \textbf{(A--B)} We show virtual exposures corresponding to a short exposure time (suffering from noise, A) and a long exposure time (suffering from motion blur, B).
    \textbf{(C)} We also display our probabilistic event representation, visualizing per-pixel temporal stability and entropy changes, where purple and green indicate per-pixel decreases and increases in entropy, respectively. \textbf{(D)} By aggregating photons till the per-pixel noise--blur limit, our demosaicked motion-aware aggregate---which includes a Wiener denoising step (\cref{fig:spatially_varying_denoising})---yields the high-fidelity color imaging.
    }
    \label{fig:color_metering}
\end{figure}

%% file: figures/analysis_exposure_time.tex
\begin{figure}[t]
    \centering
    \includegraphics[width=\columnwidth]{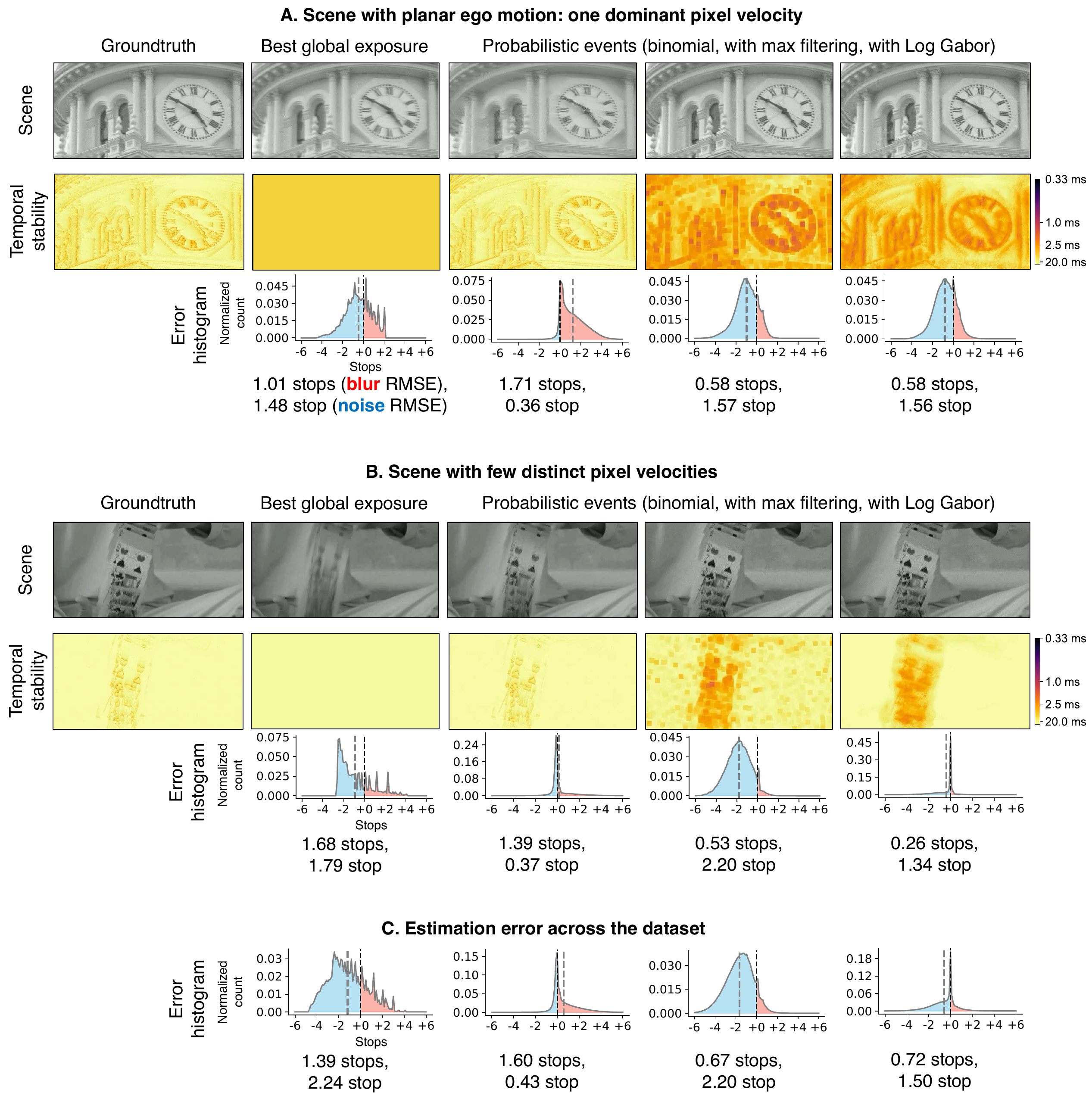}
    \vspace{-0.2in}
    \caption{\textbf{Evaluating temporal stability estimation.} We compare our probabilistic events (Log-Gabor) against the theoretical limit of a conventional sensor (Oracle Global Shutter) and ground truth. Histograms show error in stops; positive values indicate over-exposure (motion blur, marked in red), while negative values indicate under-exposure (noise, marked in blue). 
    \textbf{(A)} In scenes with uniform ego-motion, optimal integration windows are spatially correlated. The oracle global shutter approximates the ground truth reasonably well (1.01 stops blur RMSE), though our method tightens the distribution (0.58 stops blur RMSE).
    \textbf{(B)} In scenes with distinct velocities (\eg, shuffling cards), the optimal distribution is not unimodal. The global shutter is forced into a compromise, yielding high error (1.68 stops blur). Our method adapts locally, assigning short integration windows to the falling cards and long integration windows to the background, reducing blur RMSE to 0.26 stops.
    \textbf{(C)} Aggregate error across the dataset. While minimum filtering (blue dashed line) eliminates blur, it causes excessive noise (skewing left). Our Log-Gabor approach balances this trade-off, outperforming the oracle global shutter in both blur and noise metrics.}
    \label{fig:analysis_exposure_time}
\end{figure}

%% file: figures/analysis_summed_frames_vs_light_level.tex
\begin{figure}[t]
    \centering
    \includegraphics[width=\textwidth]{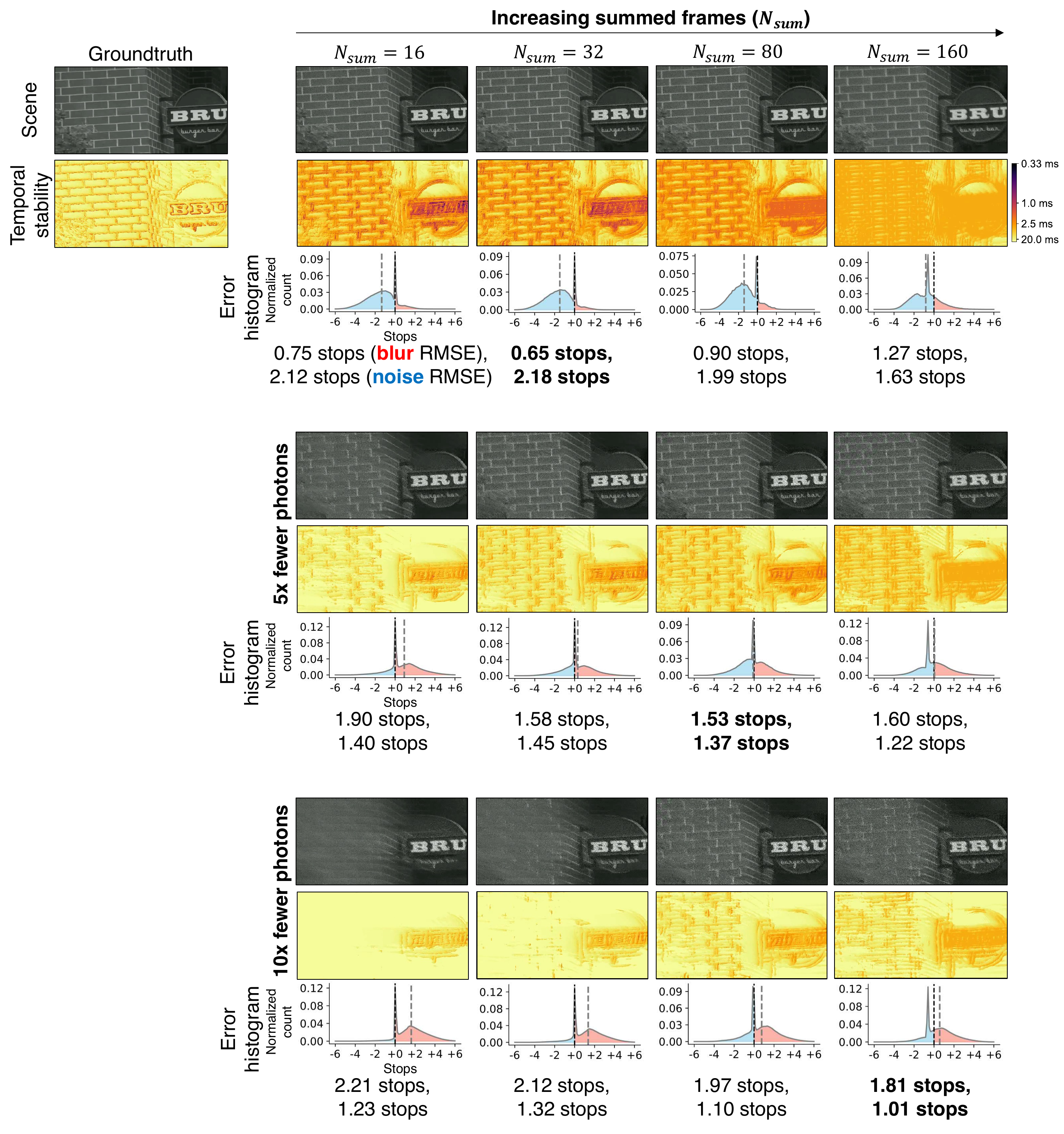}
    \vspace{-0.2in}
    \caption{\textbf{Temporal stability estimation as a function of light level and summed-frame count.}
    We visualize the effect of summed-frame count and photon rate on probabilistic-event inference for a single example sequence, complementing the aggregate error distributions in \cref{fig:analysis_summed_frames_vs_light_level_aggregate}.
    Columns vary the number of binary quanta frames summed before inference, $N_{\text{sum}}$, while rows vary the incident photon rate, from the original light level to $5\times$ fewer photons and $10\times$ fewer photons, corresponding to average photon detection rates of $0.1$ PPP and $0.05$ PPP, respectively.
    At low $N_{\text{sum}}$, estimates remain temporally responsive but are evidence-limited, especially at reduced light levels.
    Increasing $N_{\text{sum}}$ improves local evidence and stabilizes the estimate, but overly long temporal aggregation eventually introduces blur in regions with scene or camera motion.
    This example illustrates the same noise--blur trade-off observed across the dataset (shown in \cref{fig:analysis_summed_frames_vs_light_level_aggregate}): lower light requires more temporal aggregation, but the optimal summed-frame count is bounded by the temporal stability of the scene.}
    \label{fig:analysis_summed_frames_vs_light_level}
\end{figure}

%% file: figures/analysis_summed_frames_vs_light_level_aggregate.tex
\begin{figure}[t]
    \centering
    \includegraphics[width=\columnwidth]{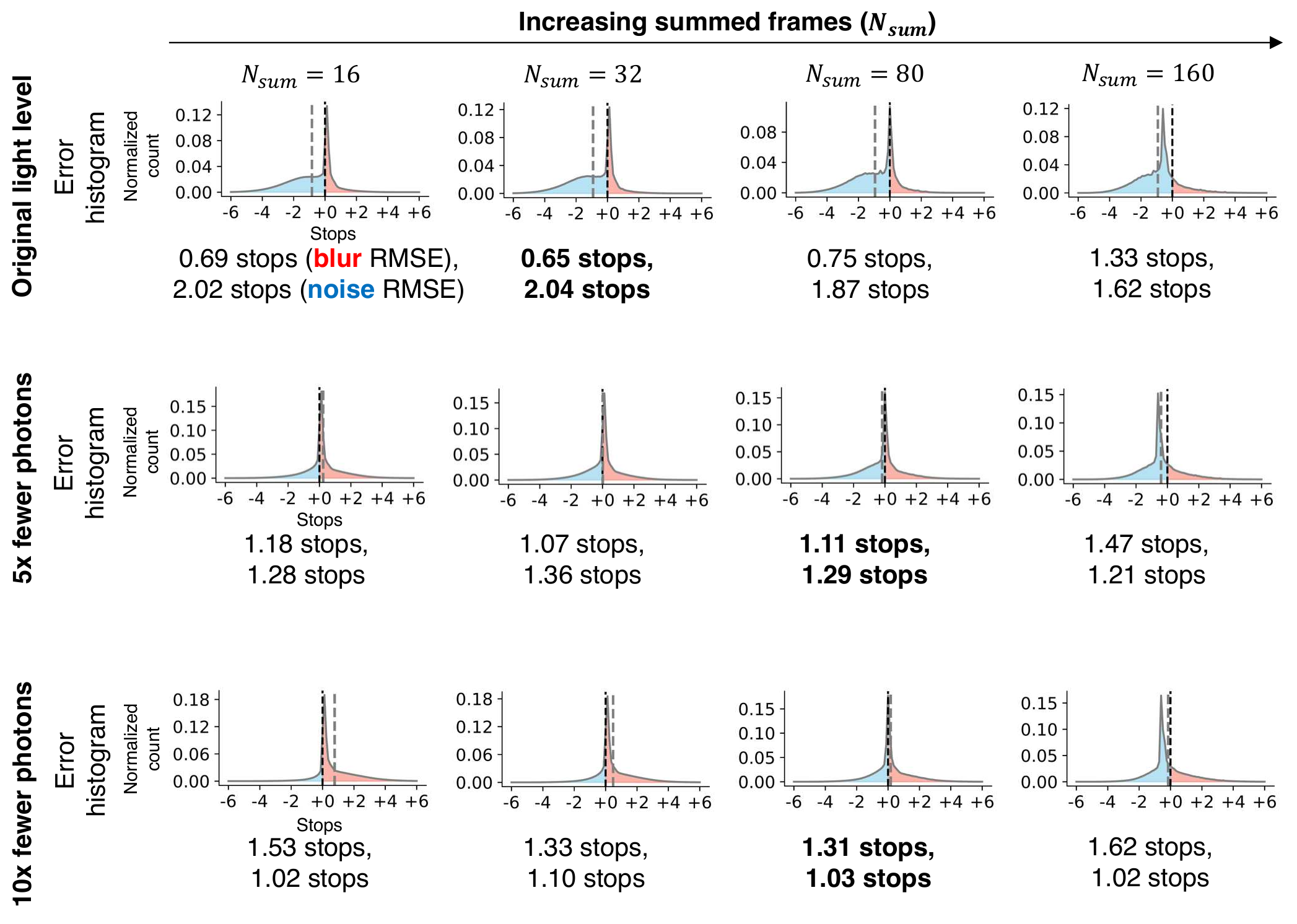}
    \vspace{-0.2in}
    \caption{\textbf{Temporal stability estimation across light levels and summed-frame counts.} We evaluate aggregate error distributions while varying the number of binary quanta frames summed before probabilistic-event inference, $N_{\text{sum}}$ (see \cref{eq:sum_image,eq:posterior_predictive_binom}), and the incident photon rate. Columns sweep $N_{\text{sum}}$ from 16 to 160, while rows show the original simulated light level, $5\times$ fewer photons, and $10\times$ fewer photons. Histograms report error in stops; positive values indicate over-exposure and motion blur (red), while negative values indicate under-exposure and noise (blue). Values below each histogram report blur RMSE and noise RMSE, respectively. As light decreases, larger summed-frame counts reduce noise but eventually increase blur, exposing the operating trade-off between photon aggregation and temporal responsiveness.}
    \label{fig:analysis_summed_frames_vs_light_level_aggregate}
\end{figure}

%% file: figures/analysis.tex
\begin{figure}[t]
    \centering
    \includegraphics[width=\columnwidth]{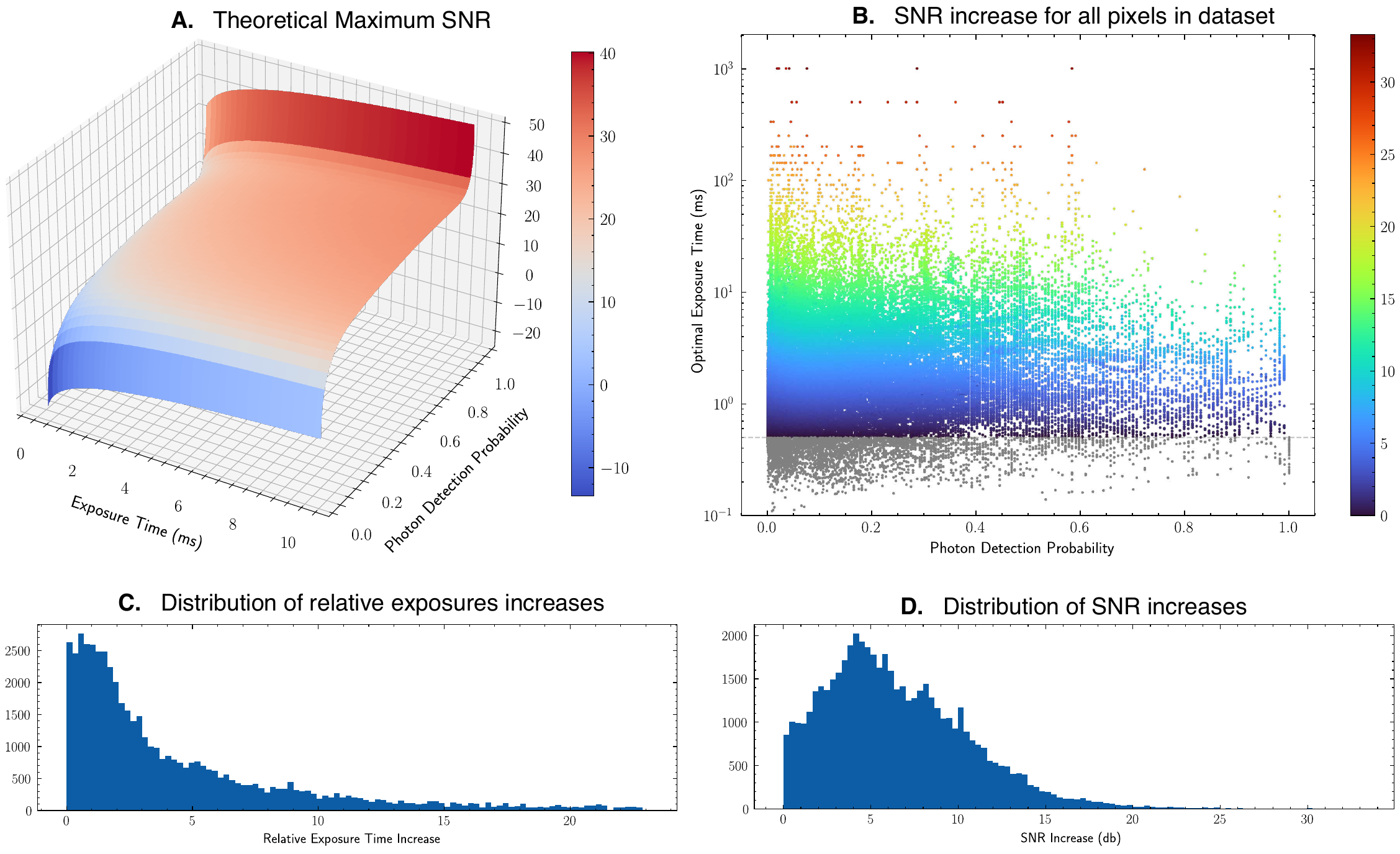}
    \vspace{-0.2in}
    \caption{\textbf{Motion-aware aggregation maximizes Signal-to-Noise Ratio (SNR) relative to global shutter baselines.} 
    \textbf{(A)} Analytical upper bound on image SNR as a function of the expected photon count per frame, $\phi$, and the integration window, $\omega$ (Eq.~\ref{eq:snr}).
    \textbf{(B)} Performance comparison on the VisionSIM dataset~\protect\citemethods{visionsim}. We plot the SNR of the optimal per-pixel integration window, $\tau_{\text{ours}}$, against the optimal global shutter baseline, $\tau_{\text{global}}$. The global exposure is constrained by the maximum scene velocity to prevent blur, with a minimum duration of \nicefrac{1}{1000}\,s. Points falling below the gray line indicate pixels with dynamics too fast for this minimum exposure, resulting in inevitable motion blur.
    \textbf{(C)} Histogram of the relative increase in integration time, $(\tau_{\text{ours}} - \tau_{\text{global}}) / \tau_{\text{global}}$, demonstrating that static background regions integrate light significantly longer than the global bottleneck permits.
    \textbf{(D)} The corresponding distribution of SNR improvement (in dB), quantifying the benefit of motion-aware aggregation.}
    \label{fig:analysis}
\end{figure}